\documentclass{article}
\def\atomsteerpreprint{}
\pdftrailerid{}
\usepackage{atomsteer_style,times}
\astfinalcopy
\makeatletter
\g@addto@macro\@maketitle{\lhead{Preprint. Under review.}}
\makeatother

\usepackage{amsmath,amssymb,amsfonts,mathtools}
\usepackage{booktabs,tabularx,multirow,array}
\usepackage{graphicx}
\usepackage{caption}
\usepackage{subcaption}
\usepackage{placeins}
\usepackage{needspace}
\usepackage{enumitem}
\usepackage{algorithm}
\usepackage{algorithmic}
\usepackage{tikz}
\usetikzlibrary{arrows.meta,positioning,fit,calc}
\usepackage{xcolor}
\usepackage{url}
\usepackage{hyperref}
\hypersetup{hidelinks}
\usepackage{natbib}
\usepackage{cleveref}
\usepackage{bm}
\usepackage{makecell}

\newcommand{\methodfull}{AtomSteer}

\newcommand{\wrap}{\operatorname{wrap}}

\usepackage{xcolor}

\title{\methodfull: Test-Time Constraint Steering for Atomistic Generative Models}

\author{Pengfei Cai\thanks{Equal contribution.}\qquad Katharina S.C. Jäger\footnotemark[1]\qquad Utkarsh Utkarsh\qquad Nofit Segal \\[2pt]
\textbf{Bowen Deng\qquad Akshay Subramanian\qquad Rafael Gomez-Bombarelli}\thanks{Corresponding author: \texttt{rafagb@mit.edu}} \\[6pt]
Massachusetts Institute of Technology
}
\date{}

\begin{document}
\maketitle
\suppressfloats[t]

\begin{abstract}
Diffusion and flow matching models offer scalable approaches to generating crystalline materials, molecular conformers, and protein-ligand complexes. However, satisfying crystallographic symmetry, structural geometry, and accurate stereochemistry remains challenging. Satisfying these requirements calls for corrections that preserve the geometry already captured by the pretrained model. In this work, we propose AtomSteer, a test-time inference framework for steering pretrained atomistic generative models with structural constraints. During sampling, AtomSteer corrects intermediate structure predictions with bounded local projections, graph-rigid stereochemical transformations, and projections over periodic symmetries, accepting updates that improve constraint satisfaction while limiting changes to the rest of the structure. We demonstrate AtomSteer across three pretrained backbones, FlowMM, ET-Flow, and Boltz-2, spanning de novo crystal generation and crystal structure prediction, molecular conformer generation, and protein-ligand complex structure prediction. AtomSteer improves space-group distribution coverage in de novo crystal generation and improves target space-group recovery in crystal structure prediction. AtomSteer improves molecular validity and \textit{E}/\textit{Z} stereochemical accuracy while retaining conformer ensemble fidelity. For protein-ligand complexes, AtomSteer reduces ligand \textit{E}/\textit{Z} errors relative to experimental reference structures while achieving competitive docking success and PoseBusters validity. Our method provides a unified framework for adding chemical and crystallographic control to existing atomistic generative models without retraining or architectural changes.

\end{abstract}

\section{Introduction}
\label{sec:introduction}

\vspace{-5pt}\begin{figure}[t]
  \centering
  \includegraphics[width=0.86\textwidth,trim=4.5 3.5 1.5 1.5,clip]{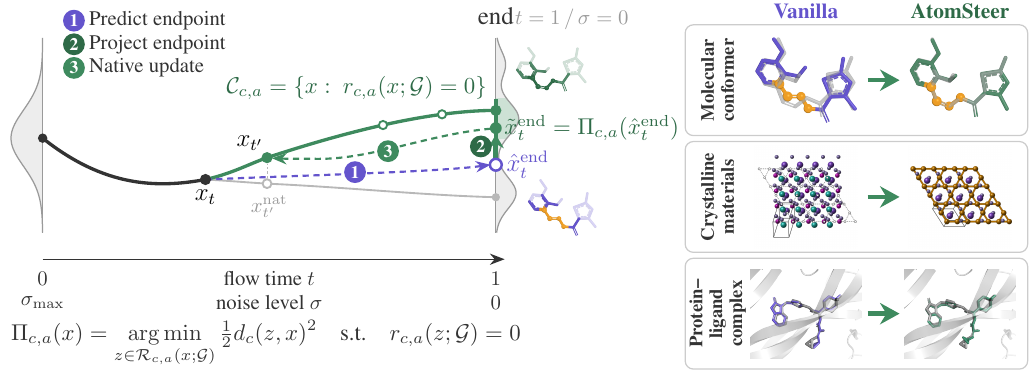}
  \caption{\textbf{AtomSteer applies structure-dependent projections during sampling.}
  (1) At state $x_t$, the pre-trained model predicts the
  endpoint $\hat x_t^{\rm end}$ (the state at $t=1$ for flows and at noise level
  $\sigma=0$ for EDM).
  (2) AtomSteer projects this prediction towards the structural constraint set
  $\mathcal C_{c,a}$ (Section~\ref{sec:atomsteer-constraints}), giving $\tilde x_t^{\rm end}$.
  (3) The native sampler continues from $\tilde x_t^{\rm end}$ to $x_{t'}$ (green); the gray line shows the unsteered trajectory.
  Fragments show an \textit{E}/\textit{Z} double bond in the uncorrected and corrected conformers.
  Right: vanilla and AtomSteer samples for a molecular conformer, a crystal with a
  requested Archimedean lattice, and a protein--ligand complex; the gray overlay shows the ground truth.}
  \label{fig:atomsteer-overview}
\vspace{-10pt}
\end{figure}

\looseness=-1
Atomistic generative models can sample periodic crystalline materials, molecular
conformers, and protein--ligand complexes with broad distributional fidelity
\citep{miller2024flowmm,hassan2024etflow,passaro2025boltz2}, yet samples often violate known structural requirements. Recent co-folding models generate wrong ligand chemistry or chirality for more than half of the tested
ligands~\citep{goddard2026badchem,ishitani2025stereo,carbohydrate}, and crystalline generative models over-sample low-symmetry structures such as $P1$
that are rare in nature~\citep{levy2025symmcd}. Each failure breaks a requirement: molecular graphs specify local geometry and stereochemistry, protein--ligand poses must be chemically valid, and space-group symmetries strongly influence the material properties~\citep{levy2025symmcd,kazeev2025wyformer}.

\looseness=-1
Existing approaches incorporate structural requirements through specialized representations, architectures, conditioning, or training objectives, while inference-time methods use reward reweighting or differentiable potentials (Section~\ref{sec:related}). A common way to add heterogeneous structural requirements at test time to pre-trained atomistic generative models is missing. These requirements differ in what they measure and in how the structure should move to satisfy them: an \textit{E}/\textit{Z} error is measured on four
atoms but corrected by rotating an entire substituent, a protein--ligand clash is measured between atom pairs but can be corrected by moving the ligand as a rigid body, and space-group symmetry is measured across symmetry-equivalent atoms but corrected by moving each orbit with the lattice. Cartesian projection, for example, is suitable for bond lengths, angles, and small torsion refinements but cannot correct an \textit{E}/\textit{Z} configuration without displacing the four dihedral atoms and distorting the bonds around them. Therefore, each structural requirement calls for a residual on the atomistic graph paired with a suitable projection.

\looseness=-1
In this work, we introduce \emph{\methodfull}, an inference-time framework that pairs the structural residual of each requirement on an atomistic graph with
structure-dependent projections (Figure~\ref{fig:atomsteer-overview}). AtomSteer projects the model's endpoint prediction at scheduled sampling steps, accepts only a bounded correction, and returns the corrected prediction through the pre-trained model's native sampling update, without retraining or modifying the model. We developed three projection families in this work, spanning a wide range of constraints: Chemical Gauss--Newton (ChemGN) for local geometry and stereochemistry, rigid-fragment and rigid-body projections for substituents, and crystal symmetry and motif projections for periodic structures. To demonstrate the framework, we test AtomSteer on FlowMM, ET-Flow, and Boltz-2 spanning the generation of crystalline materials, molecular conformers, and protein--ligand complexes (Section~\ref{sec:results}).

\looseness=-1
Our contributions are:

\begin{enumerate}[leftmargin=*,itemsep=2pt,topsep=3pt]
    \item \looseness=-1 \textbf{A test-time framework for atomistic constraints.}
    We formulate constraints as residuals on atomistic graphs and pair them with
    projection families across local geometry, stereochemistry, component
    packing, periodic symmetry, and prescribed motifs. AtomSteer applies bounded,
    structure-dependent corrections to pre-trained diffusion and flow-based
    models without retraining or architectural changes.

    \item \looseness=-1 \textbf{Crystallographic symmetry and motif control.} 
    AtomSteer infers compatible symmetry from FlowMM's
    structure predictions, broadens de novo space-group coverage (the number of distinct space groups generated), and raises joint success@1 (structure match and correct space group) from 45.18\% to
    52.54\% in MP-20 crystal structure prediction. Separately, constrained on Archimedean lattice (AL) requirements, it also achieves a higher joint rate of satisfying the AL constraint and passing SCIGEN's prescreening~\citep{okabe2026scigen}.
    
    \item \looseness=-1 \textbf{Conformer validity and ensemble quality.}
    Applied to ET-Flow, AtomSteer improves conformer precision and PoseBusters
    validity on GEOM-Drugs, reduces median \textit{E}/\textit{Z} distribution error
    by 75.9\%, and lowers GFN2-xTB ensemble-property errors by 10--36\% relative
    to the better ET-Flow vanilla sampler for each property.

    \item \looseness=-1 \textbf{Protein--ligand validity and stereochemical accuracy.}
    Applied to Boltz-2, AtomSteer achieves the highest PoseBusters validity on
    StereoBench and raises joint valid-docking-and-stereochemistry success from
    36.5\% to 45.2\% relative to Boltz-steering.
\end{enumerate}

\section{Related work}
\label{sec:related}

\vspace{-5pt}\looseness=-1
\textbf{Inference-time steering.}
Inference-time steering adapts a pre-trained diffusion or flow model by
modifying sampling rather than parameters, and it is widely used to correct the
ligands of co-folding
models~\citep{abramson2024accurate,wohlwend2025boltz1,passaro2025boltz2,chai1,protenixv1,liu2024technicalreporthelixfold3biomolecular,qiao2024neuralplexer3accuratebiomolecularcomplex,genesisresearchteam2025pearlfoundationmodelplacing}.
Particle methods such as Feynman--Kac steering reweight whole trajectories
against a reward~\citep{singhal2025fk,didi2026complexa,fksfold}. Coordinate
guidance modifies structures within single trajectories through differentiable
potentials, restraints, or terminal
optimization~\citep{ishitani2025stereo,chen2026gs,hori2026,lai2026improving,He2023ManifoldPG},
and hybrid methods combine
both~\citep{ren2026driftlitelightweightdriftcontrol,wu2024practicalasymptoticallyexactconditional,scarpellini2026fewstepcofoldingallatomflow}.
Our Boltz-steering baseline applies seven flat-bottom potentials, including chirality,
stereochemistry, internal geometry, and clashes as gradient guidance on the
denoised estimate, together with Feynman--Kac
resampling~\citep{wohlwend2025boltz1,passaro2025boltz2}. Hard projection
methods act on noisy samples~\citep{christopher2024projected} or on denoised
intermediates and predicted endpoints~\citep{utkarsh2025physics,christopher2026constraineddiffusionproteindesign}.
On protein--ligand
complexes, \citet{chen2026gs} project structures with a Gauss--Seidel solver at the end of denoising, with the denoiser fine-tuned through the
projection. AtomSteer pairs the structural residual of each atomistic graph constraint
with a structure-dependent projection whose corrections act
through local Cartesian, rigid-fragment, rigid-body, or periodic coordinates. It
projects endpoint predictions at intermediate sampling steps and returns accepted
corrections through the native update without fine-tuning, and we demonstrate across periodic crystals, molecular conformers, and protein--ligand complexes.

\looseness=-1
\textbf{Symmetry and constraints in crystal generation.}
Symmetry-aware crystal generative models build space-group or chemical constraints into
their representation, conditioning, or training: DiffCSP++ constrains diffusion
to the lattice and Wyckoff positions of a supplied space
group~\citep{jiao2024diffcsppp}, SymmCD generates asymmetric units with their
symmetry transformations~\citep{levy2025symmcd}, and CrysVCD filters
language-model compositions for charge balance before diffusion~\citep{cheng2026crysvcd}. SCIGEN imposes requested motifs at inference
by masking a diffusion model's denoising path~\citep{okabe2026scigen}. AtomSteer
instead corrects a pre-trained model's endpoint prediction at test time, inferring a
compatible symmetry without the target structure, a supplied space group, or an
energy model, and imposing requested motifs through the same projection interface. Table~\ref{tab:method_comparison} in Appendix~\ref{app:related_methods} compares the methods.

\looseness=-1
\textbf{Molecular geometry and stereochemistry.}
Graph-, fragment-, and torsion-based generative models encode geometry during training.
GeoMol predicts local structures and torsion angles, Torsional Diffusion samples
rotatable torsions, and SigmaDock diffuses rigid-fragment poses in \(SE(3)\)
\citep{ganea2021geomol,jing2022torsional,prat2025sigmadock}. Stereochemistry can
also be encoded through targeted objectives, graph augmentation, or better training
data \citep{williams2024pidm,nikitin2025loqi,lin2025moledit,vilya2026}. AtomSteer
instead projects graph-defined requirements onto the endpoint predictions of a
pre-trained model during inference.

\section{AtomSteer: Test-Time Constraint Projection}
\label{sec:method}

\vspace{-5pt}\looseness=-1
At every sampling step, the network output of a diffusion or flow model, whether expressed as noise, clean data, score, or velocity, also determines an estimate of
the clean sample toward which the current state is heading~\citep{lai2026principlesdiffusionmodels}.
Pre-trained atomistic generative models learn useful structural distributions,
but their samples are not guaranteed to satisfy chemical or
crystallographic requirements. AtomSteer adds such structural constraints to a
pre-trained atomistic sampler at test time (Figure~\ref{fig:atomsteer-overview}). Each constraint pairs a
structural residual, which defines what must be satisfied, with a
structure-dependent projection, which defines how coordinates move together to satisfy it.
AtomSteer corrects the model's \emph{endpoint prediction} $\hat x_t^{\rm end}$ of the final structure rather than the noisy state $x_t$, and the native sampler then updates the projected endpoint $\tilde x_t^{\rm end}$ to the next state $x_{t'}$, so later
sampling steps respond to the corrections (Figure~\ref{fig:atomsteer-overview}).

\subsection{Structural residuals and structure-dependent projections}
\label{sec:atomsteer-constraints}

\vspace{-5pt}\looseness=-1
Let $\mathcal G=(V,\mathcal E,\xi)$ be an atomistic graph, where its vertices $V$ are atoms or crystal sites. Its relations $\mathcal E$ group the atoms that a constraint involves, such as a bonded pair, the four atoms around a double bond, or the designated sites of a crystal motif. The labels $\xi$ carry the properties, such as bond orders, stereochemical labels, and, when fixed, atomic species. The model generates the configuration $x$, such as coordinates, lattice parameters, or atom types.
For a constraint $c$ and a discrete realization $a$, such as one space-group setting, we define a violation residual and its constraint set by
\begin{equation}
 r_{c,a}(x;\mathcal G),\qquad
 \mathcal C_{c,a}=\{x:r_{c,a}(x;\mathcal G)=0\}.
 \label{eq:atomsteer-realizations}
\end{equation}
For example, a steric residual measures how far two atoms fall below their minimum separation, and a periodic-symmetry residual measures the disagreement between an atom and its symmetry-related structure.

\looseness=-1
A structure-dependent projection reduces these violations by moving the appropriate coordinates together. With $\mathcal R_{c,a}(x;\mathcal G)$ the structures reachable through these moves, it seeks the constraint-satisfying structure with minimum displacement,
\begin{equation}
 \Pi_{c,a}(x)=
 \operatorname*{arg\,min}_{z\in\mathcal R_{c,a}(x;\mathcal G)}
 \frac12 d_c(z,x)^2
 \quad\text{subject to}\quad r_{c,a}(z;\mathcal G)=0,
 \label{eq:atomsteer-minimum-intervention}
\end{equation}
within the projection's movement budget. Bounded numerical corrections take damped
steps on the corresponding residual--displacement objective,
\begin{equation}
 \min_{z\in\mathcal R_{c,a}(x;\mathcal G)}
 \frac12\|r_{c,a}(z;\mathcal G)\|_{W_c}^2
 +\frac{\lambda_c}{2}d_c(z,x)^2.
 \label{eq:atomsteer-relaxed-projection}
\end{equation}
The distance $d_c$ measures structural displacement, $W_c$ weights the residuals, and $\lambda_c\ge 0$ controls the displacement penalty. Setting $\lambda_c=0$ removes this penalty, so only the movement budget limits the displacement. 

Table~\ref{tab:atomsteer-instantiations} organizes the three projection families that realize these objectives by the coordinates each corrects. These projections instantiate one interface: each supplies a residual and a structure-dependent projection, and all
share the same endpoint intervention, bounded acceptance, and native update.
On ET-Flow and Boltz-2 the molecular projections use the same implementation.

\looseness=-1
Generally, molecular stereochemistry and requested crystal motifs specify their targets directly. When a crystal's space group is not provided, AtomSteer infers it from the model's own endpoint prediction through hierarchy-aware symmetry
selection. This selection enumerates the space groups that the endpoint prediction already approximately satisfies and promotes along
minimal-supergroup relations to the most symmetric feasible parent group.
AtomSteer then projects the atomic positions onto symmetry orbits of that group in fractional coordinates, and the lattice metric onto the group's metric constraints, with atom types fixed
(Appendix~\ref{app:periodic-residuals}).

\begin{table}[t]
\centering
\caption{\textbf{Structural constraints grouped by projection family.}
One requirement can combine families: e.g., \textit{E}/\textit{Z} fragment flips and
fragment refinement are followed by \textit{E}/\textit{Z} ChemGN (Cartesian). Appendices~\ref{app:shared-projections}, \ref{app:sampling},
and~\ref{app:protein-ligand-methods} define the residuals and details.}
\label{tab:atomsteer-instantiations}
\small
\setlength{\tabcolsep}{3pt}
\renewcommand{\arraystretch}{1.0}
\begin{tabularx}{\textwidth}{>{\raggedright\arraybackslash}p{.165\textwidth}>{\raggedright\arraybackslash}p{.335\textwidth}>{\raggedright\arraybackslash}p{.215\textwidth}>{\raggedright\arraybackslash}X}
\toprule
Projection family & Structural constraint examples & Coordinates corrected & Projection \\
\midrule
Chemical Gauss--Newton (ChemGN)
 & Local chemical geometry and sterics: bond and 1--3 geometry, designated planarity, intramolecular clashes, protein--ligand exclusion, and \textit{E}/\textit{Z} dihedral error
 & Selected molecular or ligand coordinates
 & Cartesian ChemGN \\
 & Local \textit{R}/\textit{S} handedness
 & Ligand coordinates with first-order bond protection
 & Bond-tangent ChemGN \\
\midrule
Rigid-fragment and rigid-body projections
 & \textit{E}/\textit{Z} configuration
 & Connected substituent around the double bond
 & \textit{E}/\textit{Z} fragment flips and refinement \\
 & Component packing
 & Ligand or eligible protein component
 & Rigid-body $SE(3)$ \\
\midrule
Crystal symmetry and motif projections
 & Space-group symmetry
 & Fractional coordinates and lattice
 & Hierarchy-aware symmetry selection, orbit projection, and lattice-metric projection \\
 & Lattice-metric relations
 & Lattice
 & Metric-template projection \\
 & Requested Archimedean lattice
 & Selected species, fractional positions, and lattice entries
 & Motif projection \\
\bottomrule
\end{tabularx}
\vspace{-10pt}
\end{table}

\subsection{Endpoint projection during sampling}
\label{sec:atomsteer-core}

\vspace{-5pt}\looseness=-1
The three pre-trained backbones span Cartesian flow matching, Riemannian flow
matching, and EDM-style denoising diffusion. Their endpoint predictions are
\begin{equation}
 \hat x_t^{\rm end}=\begin{cases}
 x_t+(1-t)v_\theta(x_t,t),&\text{ET-Flow},\\
 \operatorname{projx}[x_t+(1-t)v_\theta(x_t,t)],&\text{FlowMM},\\
 D_\theta(x_\sigma,\sigma),&\text{Boltz-2}.
 \end{cases}
 \label{eq:atomsteer-endpoint-models}
\end{equation}
The flow models extrapolate the raw learned velocity $v_\theta$ to $t=1$, with
$\operatorname{projx}$ applying FlowMM's representation operations, including
periodic wrapping. Boltz-2 follows EDM, in which $\sigma$ is the noise level of unscaled noisy
coordinates $x_\sigma=x_0+\sigma\epsilon$ with $\epsilon\sim\mathcal N(0,I)$, and its denoiser $D_\theta$ maps
$x_\sigma$ to an estimate of the clean ($\sigma=0$)
coordinates $x_0$~\citep{karras2022edm,passaro2025boltz2}. Its step from $t$ to $t'$
below is one EDM update (Appendix~\ref{app:protein-ligand-methods}).

\looseness=-1
At scheduled steps, the projected endpoint replaces the endpoint prediction in
the model's native update from $x_t$ to $x_{t'}$. Each projection bounds its
correction, because fully satisfying one constraint can distort nearby
geometry, such as the bonds next to a removed steric clash. A
correction that fails its acceptance checks is discarded, and those
coordinates keep their native update from the pre-trained model. For more details, Appendices~\ref{app:crystal-methods}, \ref{app:conformer-methods},
and~\ref{app:protein-ligand-methods} give each backbone's instantiation and
schedule, and Appendix~\ref{app:shared-projections}
defines the shared molecular projections.

\section{Results}
\label{sec:results}
\suppressfloats[t]

\vspace{-5pt}\looseness=-1
To demonstrate, we apply AtomSteer to three pre-trained models with a broad range of structural constraint types
(Section~\ref{sec:atomsteer-core}). For each experiment, we ask whether the
targeted constraint is better satisfied and whether a scientifically relevant downstream outcome improves with it (protocols elaborated in Appendices~\ref{app:crystalline-materials},
\ref{app:molecular-conformers}, and~\ref{app:protein-ligand}).

\subsection{AtomSteer improves crystallographic symmetry in FlowMM}
\label{sec:results-flowmm}

\vspace{-5pt}\looseness=-1
On FlowMM~\citep{miller2024flowmm}, we look into tasks from enforcing global symmetry in MP-20~\citep{Horton_2025} \textit{de novo} generation, target symmetry in crystal structure prediction (CSP) on MP-20, and constraining requested local motifs~\citep{okabe2026scigen}.

\looseness=-1
\paragraph{De novo crystal generation.}
\label{sec:results-flowmm-denovo}

\begin{figure*}[t]
    \centering
    \includegraphics[width=0.8\textwidth]{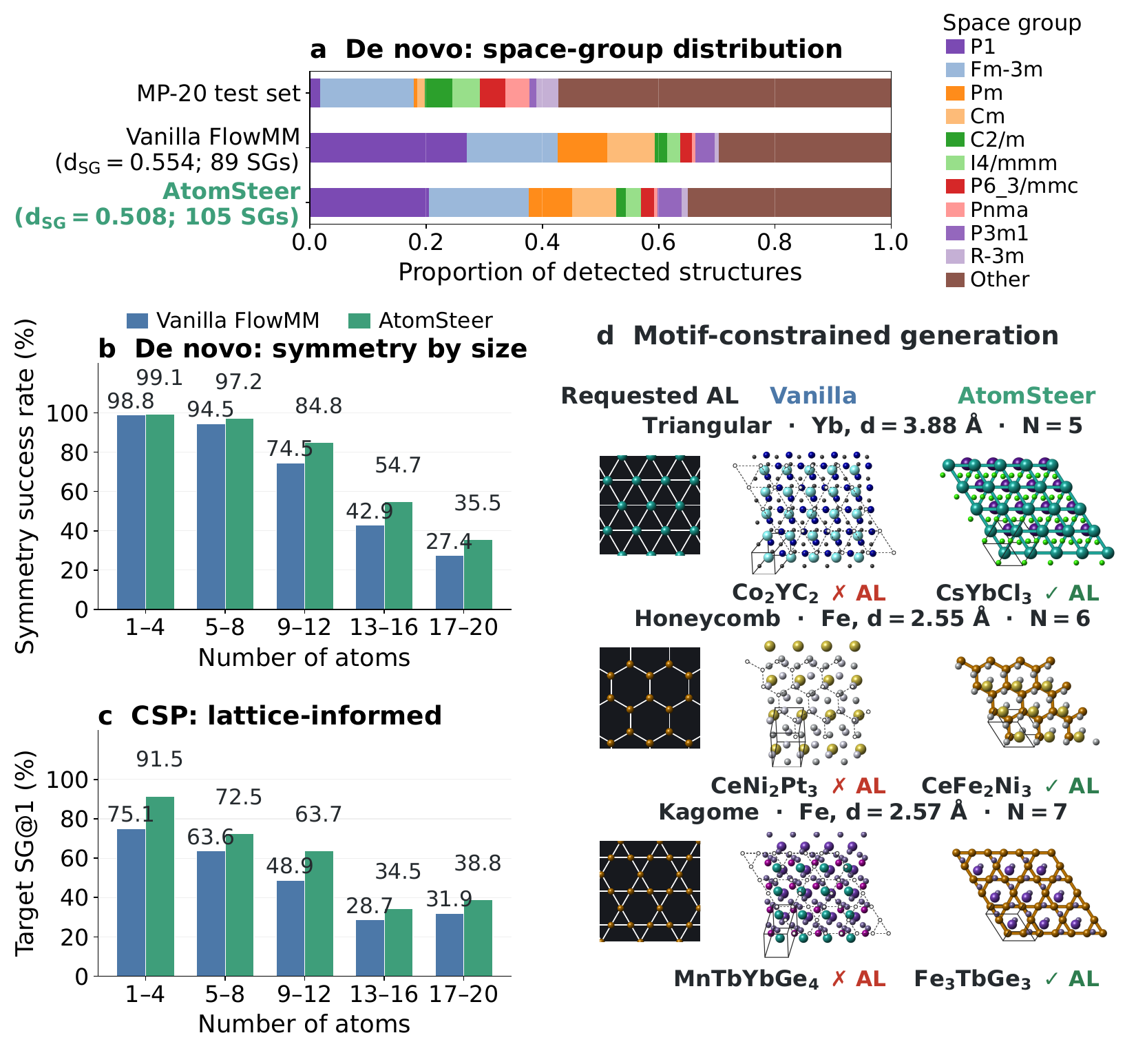}
    \caption{\textbf{Crystallographic symmetry and requested structures in FlowMM.}
    \textbf{a,} Detected space-group distributions of the MP-20 test set,
    vanilla FlowMM, and AtomSteer in \textit{de novo} generation.
    \textbf{b,} Symmetry success rate of paired \textit{de novo} samples by
    number of atoms in the generated unit cell.
    \textbf{c,} Target-space-group recovery (SG@1) for lattice-informed
    MP-20 CSP by number of atoms.
    \textbf{d,} Requested triangular, honeycomb, and Kagome Archimedean
    lattices (ALs) beside matched vanilla FlowMM and AtomSteer structures;
    dashed guides in the vanilla panels mark where AtomSteer realizes the AL.
    Metrics, evaluation subsets, and tolerances: Appendix~\ref{app:crystal-metrics}.}
    \label{fig:flowmm-symmetry-overview}
\vspace{-3pt}
\end{figure*}

\begin{table}[!htbp]
\centering
\caption{\textbf{Crystallographic symmetry in MP-20 \textit{de novo} generation.}
Each method uses 10{,}000 paired samples. Symmetry success is the fraction of all samples assigned
a space-group number $\geq 3$; $d_{\mathrm{SG}}$ is the Jensen--Shannon distance to
the MP-20 space-group distribution; Unique SGs counts distinct detected groups; and Valid measures as-generated geometric
validity. $\Delta F_{\max}$ is the median paired change in maximum atomic force
relative to FlowMM. Relaxed columns use full-cell UMA
relaxation~\citep{wood2025uma}. Best results are \textbf{bolded}, and second-best \underline{underlined}. Details: Appendices~\ref{app:crystal-methods},
\ref{app:crystal-denovo-protocol}, and~\ref{app:crystal-additional-results}.}
\label{tab:flowmm-denovo}
\small
\setlength{\tabcolsep}{2.2pt}
\begin{tabular*}{\textwidth}{@{\extracolsep{\fill}}lrrrrrrr@{}}
\toprule
& \multicolumn{5}{c}{\textit{As generated}} & \multicolumn{2}{c}{\textit{After UMA relaxation}} \\
\cmidrule(lr){2-6}\cmidrule(lr){7-8}
Method
& \makecell{Symmetry success\\rate (\%) $\uparrow$}
& $d_{\mathrm{SG}}$ $\downarrow$
& \makecell{Unique\\SGs $\uparrow$}
& \makecell{Valid\\(\%) $\uparrow$}
& \makecell{Paired $\Delta F_{\max}$\\(eV/\AA) $\downarrow$}
& \makecell{Symmetry success\\rate (\%) $\uparrow$}
& $d_{\mathrm{SG}}$ $\downarrow$ \\
\midrule
FlowMM
& \underline{71.47} & \underline{0.5541} & \underline{89}
& \textbf{97.82} & -- & \underline{64.19} & \underline{0.5363} \\
FlowMM + AtomSteer
& \textbf{77.91} & \textbf{0.5083} & \textbf{105}
& \underline{97.79} & $-0.0394$ & \textbf{71.50} & \textbf{0.5095} \\
\bottomrule
\end{tabular*}
\vspace{-3pt}
\end{table}

\looseness=-1
We ask whether
AtomSteer can move FlowMM's space-group distribution toward MP-20 without a
supplied target group.
Hierarchy-aware symmetry selection promotes the symmetry that FlowMM's
predicted structure already approximates to the most symmetric feasible parent
group and projects the atomic symmetry orbits and lattice metric onto it
(Section~\ref{sec:atomsteer-constraints}).

\looseness=-1
\textbf{AtomSteer moves the de novo generated space-group distribution toward MP-20,
especially with improvements for cells with more atoms.}
AtomSteer raises the symmetry success rate from 71.47\% to 77.91\%, broadens
the generated space-group support, and lowers $d_{\mathrm{SG}}$
(Table~\ref{tab:flowmm-denovo}; Figure~\ref{fig:flowmm-symmetry-overview}a).
The gain is clearer in unit cells with more atoms, where vanilla FlowMM most
often loses symmetry (Figure~\ref{fig:flowmm-symmetry-overview}b), and persists after
full-cell UMA relaxation (Table~\ref{tab:flowmm-denovo};
Appendix~\ref{app:crystal-additional-results}). AtomSteer therefore recovers
symmetry that the pre-trained trajectory already approaches, moving the generated
space-group distribution toward MP-20 without a supplied target group.

\looseness=-1
\paragraph{Crystal structure prediction.}
\label{sec:results-flowmm-csp}

\begin{table}[!htbp]
\centering
\caption{\textbf{Crystal structure prediction on MP-20 ($N=9{,}046$).}
Values (\%) measure structure match (MR@K), target-space-group recovery (SG@K),
and both on one candidate (Joint@K) within $K$ samples. $\dagger$ marks published
MR values~\citep{xie2022cdvae,jiao2023diffcsp,luo2025crystalflow,hollmer2025omatg,seong2026mcflow,miller2024flowmm};
other rows use the common evaluator. AtomSteer (lattice-informed) also receives a
reference-derived lattice metric template and integer setting; other rows receive
composition only. Best \textbf{bold}, second-best \underline{underlined}.
Evaluation details: Appendix~\ref{app:crystal-csp-protocol}.}
\label{tab:flowmm-csp}
\small
\setlength{\tabcolsep}{4.3pt}
\renewcommand{\arraystretch}{0.93}
\begin{tabular*}{\textwidth}{@{\extracolsep{\fill}}lrrrrrr@{}}
\toprule
& \multicolumn{2}{c}{MR $\uparrow$}
& \multicolumn{2}{c}{SG $\uparrow$}
& \multicolumn{2}{c}{Joint $\uparrow$} \\
\cmidrule(lr){2-3}\cmidrule(lr){4-5}\cmidrule(lr){6-7}
Method & @1 & @20 & @1 & @20 & @1 & @20 \\
\midrule
CDVAE$^{\dagger}$
    & 33.90 & 66.95 & -- & -- & -- & -- \\
DiffCSP
    & 52.21 & 77.77 & 47.56 & 75.41 & 41.50 & 65.49 \\
CrystalFlow
    & 62.15 & \underline{78.27} & 3.06 & 15.03 & 1.64 & 12.77 \\
OMatG
    & 63.50 & 74.38 & 1.86 & 2.30 & 0.63 & 1.35 \\
MCFlow-L$^{\dagger}$
    & \underline{64.08} & 76.08 & -- & -- & -- & -- \\
FlowMM
    & 62.45 & 75.29 & 51.66 & 71.17 & 45.18 & 62.44 \\
FlowMM + AtomSteer (composition-only)
    & 62.26 & 74.94 & \underline{59.83} & \underline{75.81} & \underline{52.54} & \underline{66.32} \\
FlowMM + AtomSteer (lattice-informed)
    & \textbf{66.33} & \textbf{81.65} & \textbf{62.68} & \textbf{80.69} & \textbf{54.95} & \textbf{70.61} \\
\bottomrule
\end{tabular*}
\vspace{-3pt}
\end{table}

\looseness=-1
For CSP, we ask whether AtomSteer can recover the target symmetry together with
the crystal structure of a given composition. \textbf{Structure match alone can
hide crystallographic errors.} Benchmarks usually score CSP by the
StructureMatcher match rate (MR@K)~\citep{jiao2023diffcsp,miller2024flowmm}, and often omit space group accuracy, so we also report target-space-group recovery (SG@K) and Joint@K, which requires one candidate to satisfy both. Note that CrystalFlow~\citep{luo2025crystalflow}
and OMatG~\citep{hollmer2025omatg} reach MR@1 on par with FlowMM, yet attain SG@1 of only 3.06\% and 1.86\% (Table~\ref{tab:flowmm-csp}).

\looseness=-1
\textbf{AtomSteer (composition-only) improves space-group and joint recovery while preserving match rate.}
With the symmetry inferred as in \textit{de novo} generation, AtomSteer raises FlowMM's MP-20 Joint@1 from 45.18\% to 52.54\% while MR@1 is essentially unchanged, with the highest SG and Joint values
among composition-only methods at $K=1$ and 20 (Table~\ref{tab:flowmm-csp}). Per-size results are in
Appendix~\ref{app:crystal-additional-results}.

\looseness=-1
\textbf{Lattice-system information further improves structure and joint recovery.}
Powder diffraction typically determines the crystal system before the atomic structure is solved~\citep{park2017crystal,suzuki2020symmetry}. AtomSteer
(lattice-informed) receives the corresponding discrete information from the reference crystal: its lattice system, as a metric template, and the integer transform from FlowMM's cell to the conventional cell. The pre-trained FlowMM still generates the numerical cell parameters and atomic positions, and hierarchy-aware symmetry selection infers the space group from the generated structure
(Appendix~\ref{app:periodic-residuals}). AtomSteer attains the highest performance in every column of Table~\ref{tab:flowmm-csp} (SG@1 by size in Figure~\ref{fig:flowmm-symmetry-overview}c), including 66.33\% MR@1 and 54.95\%
Joint@1. In both regimes, AtomSteer improves crystallographic correctness without sacrificing FlowMM's structure match.

\looseness=-1
\paragraph{Motif-constrained generation.}
\label{sec:results-flowmm-scigen}

\begin{table}[!htbp]
\centering
\caption{\textbf{Motif-constrained generation with AtomSteer on FlowMM against SCIGEN.}
Each row uses all 1{,}000 planned conditions for its Archimedean lattice (AL),
scored by the common evaluator. Joint requires AL constraint satisfaction and
prescreening survival on one sample; AL detection after relaxation requires a
valid, converged UMA-relaxed structure. Best within each AL and metric \textbf{bold},
second-best \underline{underlined}. Protocols: Appendices~\ref{app:crystal-motif-protocol}
and~\ref{app:crystal-methods}.}
\label{tab:flowmm-scigen}
\small
\setlength{\tabcolsep}{5pt}
\renewcommand{\arraystretch}{0.93}
\begin{tabular}{@{}llrrrr@{}}
\toprule
AL & Method & \makecell{AL constraint\\satisfied (\%) $\uparrow$} & \makecell{Prescreening\\survival (\%) $\uparrow$} & \makecell{Joint\\(\%) $\uparrow$} & \makecell{AL detected after\\relaxation (\%) $\uparrow$} \\
\midrule
\multirow{3}{*}{Kagome}
 & SCIGEN & \underline{98.3} & 8.0 & \underline{8.0} & \underline{10.3} \\
 & FlowMM (vanilla) & 0.1 & \textbf{45.2} & 0.1 & 0.3 \\
 & AtomSteer & \textbf{98.8} & \underline{11.4} & \textbf{10.9} & \textbf{10.5} \\
\midrule
\multirow{3}{*}{Honeycomb}
 & SCIGEN & \underline{98.4} & 7.4 & \underline{7.4} & \textbf{13.1} \\
 & FlowMM (vanilla) & 0.0 & \textbf{47.3} & 0.0 & 0.0 \\
 & AtomSteer & \textbf{99.2} & \underline{8.5} & \textbf{7.7} & \underline{9.4} \\
\midrule
\multirow{3}{*}{Triangular}
 & SCIGEN & \textbf{97.5} & 2.5 & \underline{2.5} & \underline{31.5} \\
 & FlowMM (vanilla) & 0.1 & \textbf{61.4} & 0.1 & 2.5 \\
 & AtomSteer & \underline{96.6} & \underline{33.1} & \textbf{30.7} & \textbf{86.4} \\
\bottomrule
\end{tabular}
\vspace{-4pt}
\end{table}

\looseness=-1
Magnetic atoms on triangular, honeycomb, or Kagome lattices can host exotic magnetic states such as quantum spin liquids~\citep{okabe2026scigen}. We therefore ask whether AtomSteer can impose on FlowMM one of these three primary Archimedean
lattices (ALs), with a specified magnetic element on its vertices, bond length, and atom count drawn as in SCIGEN, and screen stability with SCIGEN's released four-stage prescreening
filter (Appendix~\ref{app:crystal-motif-protocol}).
AtomSteer projects FlowMM's predicted structure onto the requested AL by
setting the vertex species, vertex positions, and in-plane lattice
(Appendix~\ref{app:motif-residuals}).

\looseness=-1
\textbf{AtomSteer turns unconstrained FlowMM into an AL-constrained generator.}
Vanilla FlowMM inference almost never produces the requested AL, whereas AtomSteer
realizes it in 96.6--99.2\% of samples, on par with SCIGEN
(Table~\ref{tab:flowmm-scigen}; Figures~\ref{fig:flowmm-symmetry-overview}d
and~\ref{fig:flowmm-motif-examples}). SCIGEN is designed specifically for this task and imposes the requested AL at every denoising step of its DiffCSP model. AtomSteer also attains the highest joint rate (AL constraint satisfied and prescreening passed)
for all three ALs, and the requested triangular AL is detected after relaxation in
86.4\% of its samples, against 31.5\% for SCIGEN. This demonstrates that the same framework extends from inferred global symmetry to explicitly requested local motifs in materials generation.

\FloatBarrier
\subsection{AtomSteer improves conformer validity, quality, and stereochemistry in ET-Flow}
\label{sec:results-etflow}

\begin{table}[!htbp]
\centering
\caption{\textbf{Conformer generation and GFN2-xTB ensemble properties on GEOM-Drugs.} \textbf{a,} Equal-molecule averages
with $2K$ generated conformers for $K$ references. COV (0.75~\AA{} threshold) and
AMR (\AA{}) measure coverage and mean minimum RMSD over references (R) or generated
conformers (P). PB-valid is the PoseBusters validity rate of the first $\min(2K,32)$ conformers.
\textit{E}/\textit{Z} $W_1$ is the Wasserstein distance to GEOM dihedral distributions,
median over 153 molecules with an evaluable stereogenic double bond.
\textbf{b,} Median absolute errors of Boltzmann-weighted GFN2-xTB~\citep{bannwarth2019gfn2xtb}
ensemble properties (mean energy $E$, dipole moment $\mu$, HOMO--LUMO gap
$\Delta\epsilon$, minimum energy $E_{\min}$) on a fixed random 100-molecule panel.
ET-Flow and ET-Flow-SS are our reruns of the released deterministic and stochastic 50-step samplers of \citet{hassan2024etflow};
AtomSteer uses the same pre-trained checkpoint. Best \textbf{bold}, second-best
\underline{underlined}. Protocols: Appendices~\ref{app:conformer-metrics}
and~\ref{app:conformer-methods}.}
\label{tab:etflow-main}

\footnotesize
\setlength{\tabcolsep}{2.6pt}
\resizebox{\textwidth}{!}{%
\begin{tabular}{@{}lrrrrrrrrrr@{}}
\toprule
& \multicolumn{6}{c}{\textbf{a. Conformer generation}}
& \multicolumn{4}{c}{\textbf{b. GFN2-xTB ensemble properties}} \\
\cmidrule(lr){2-7}\cmidrule(lr){8-11}
Method
& COV-R $\uparrow$
& COV-P $\uparrow$
& AMR-R $\downarrow$
& AMR-P $\downarrow$
& PB-valid $\uparrow$
& \shortstack{\textit{E}/\textit{Z} $W_1$ $\downarrow$\\($^\circ$)}
& \shortstack{$E$ $\downarrow$\\(kcal/mol)}
& \shortstack{$\mu$ $\downarrow$\\(D)}
& \shortstack{$\Delta\epsilon$ $\downarrow$\\(kcal/mol)}
& \shortstack{$E_{\min}$ $\downarrow$\\(kcal/mol)} \\
\midrule
ET-Flow
& \underline{79.92}
& 73.40
& 0.458
& 0.562
& 93.13
& 9.04
& \underline{0.1795}
& 0.2245
& \underline{0.3397}
& 0.0217 \\
ET-Flow-SS
& \textbf{79.93}
& \underline{73.52}
& \textbf{0.456}
& \underline{0.559}
& \underline{93.17}
& \underline{8.94}
& 0.1812
& \underline{0.2193}
& 0.3413
& \underline{0.0197} \\
\textbf{AtomSteer}
& 79.01
& \textbf{75.49}
& \underline{0.457}
& \textbf{0.529}
& \textbf{96.93}
& \textbf{2.15}
& \textbf{0.1610}
& \textbf{0.1558}
& \textbf{0.2354}
& \textbf{0.0126} \\
\bottomrule
\end{tabular}}

\end{table}

\looseness=-1
We ask whether enforcing the local geometry and stereochemistry also improves ET-Flow's conformer ensembles and their quantum-chemical properties. AtomSteer applies rigid-fragment \textit{E}/\textit{Z} projections, ChemGN, and a terminal projection to ET-Flow on the GEOM-Drugs test set~\citep{axelrod2022geom}
(Appendix~\ref{app:conformer-methods}).

\looseness=-1
\textbf{AtomSteer improves stereochemistry, validity, and conformer precision
while retaining ensemble coverage.}
Relative to the stronger ET-Flow sampler, ET-Flow-SS, AtomSteer lowers the median \textit{E}/\textit{Z}
distribution error by 75.9\%, raises PoseBusters validity to 96.93\%, and improves COV-P and AMR-P, while COV-R stays at 79.01\% against 79.93\% (Table~\ref{tab:etflow-main}a). Appendix
Figures~\ref{fig:etflow-fixes} and~\ref{fig:etflow-conformers} showcase some visual examples of constraint corrections and conformer samples.

\looseness=-1
\textbf{The geometry corrections also improve quantum-chemical ensemble
properties.}
After relaxation with GFN2-xTB, AtomSteer attains
the lowest median error on all four ensemble properties (as evaluated in the original ET-Flow paper), 10--36\%
below the better ET-Flow sampler for each property
(Table~\ref{tab:etflow-main}b; Appendix~\ref{app:conformer-xtb-protocol}). These gains show that correcting conformer geometry can improve the physical accuracy of the ensembles.

\subsection{AtomSteer improves \textit{E}/\textit{Z} geometry and physical validity in Boltz-2}
\label{sec:results-boltz}

\looseness=-1
Co-folding models receive ligand stereochemistry in the input graph, yet their predicted complexes often invert or distort it. Existing steering targets physical validity and stereochemical descriptors. We therefore test whether AtomSteer can also correct ligand geometry and improve double bond stereochemistry on Boltz-2~\citep{passaro2025boltz2}. AtomSteer enforces these graph-derived constraints with rigid-fragment \textit{E}/\textit{Z} projections, ChemGN, and rigid-body projections (Section~\ref{sec:atomsteer-constraints}).

\looseness=-1
We compare against unsteered Boltz-2 (Vanilla) and Boltz-steering on the PoseBusters benchmark~\citep{buttenschoen2024posebusters} and StereoBench. The latter comprises 1{,}103 X-ray protein--ligand complexes with at least one stereogenic double bond per ligand, allowing a closer evaluation of \textit{E}/\textit{Z} isomerism. On StereoBench we also evaluate Boltz-steering's potential guidance without its Feynman--Kac resampling (Guidance only), the stereochemical restraint guidance of \citet{ishitani2025stereo}, and the inference-only Gauss--Seidel projection of \citet{chen2026gs} on Boltz-2.
One primary \textit{E}/\textit{Z} metric is the fraction of ligands whose every stereogenic double bond lies within $10^\circ$ of the experimental dihedral. The discrete \textit{E}/\textit{Z} descriptor, which the PoseBusters double bond check compares, is saturated at $99.4$--$99.6\%$ for most steering methods and does not measure how far the dihedral deviates from the experimental structure.
Metric definitions and evaluation subsets are in Appendix~\ref{app:protein-ligand-metrics}, and benchmark information and baseline settings in Appendices~\ref{app:protein-ligand-tasks} and~\ref{app:protein-ligand-methods}.

\begin{table}[!htb]
\centering
\caption{\textbf{Protein--ligand steering on the PoseBusters benchmark (301 targets) and
StereoBench (1{,}103 targets).} Entries are percentages except median \textit{E}/\textit{Z}
error. Methods apply inference-time steering to the same pre-trained Boltz-2.
Docking success requires RMSD $<2$~\AA{}; Success@1 and Success@5 require a
PB-valid pose with RMSD $<2$~\AA{} (confidence top-1; any of five). Best \textbf{bold}, second-best
\underline{underlined}. Definitions: Appendix~\ref{app:protein-ligand-metrics}.}
\label{tab:boltz-final}
\setlength{\tabcolsep}{4pt}
\renewcommand{\arraystretch}{0.93}
\resizebox{\textwidth}{!}{%
\begin{tabular}{l ccc @{\hspace{1.6em}} cccccc}
\toprule
& \multicolumn{3}{c}{\textbf{PoseBusters benchmark}} & \multicolumn{6}{c}{\textbf{StereoBench}} \\
\cmidrule(lr){2-4} \cmidrule(lr){5-10}
Metric & Vanilla & \shortstack{Boltz-\\steering} & \textbf{AtomSteer} & Vanilla & \shortstack{Guidance\\only} & \shortstack{Boltz-\\steering} & \shortstack{Ishitani \&\\Moriwaki} & \shortstack{Chen\\et al.} & \textbf{AtomSteer} \\
\midrule
PB-valid $\uparrow$ & \underline{79.1} & \textbf{99.3} & \textbf{99.3} & 51.0 & 90.7 & \underline{91.7} & 67.6 & 87.1 & \textbf{93.4} \\
RMSD $<2$~\AA{} $\uparrow$ & 83.7 & \underline{84.1} & \textbf{84.7} & 63.2 & 64.0 & \underline{64.5} & 62.2 & 64.2 & \textbf{64.7} \\
Success@1 $\uparrow$ & 66.1 & \underline{83.4} & \textbf{84.1} & 35.7 & 59.1 & \underline{59.7} & 45.7 & 56.8 & \textbf{60.7} \\
Success@5 $\uparrow$ & 78.3 & \underline{87.3} & \textbf{88.7} & 46.9 & \underline{66.0} & 65.8 & 57.9 & 63.3 & \textbf{67.5} \\
\textit{E}/\textit{Z} median error ($^\circ$) $\downarrow$ & 6.12 & \underline{5.26} & \textbf{2.98} & 6.27 & 6.44 & \underline{6.13} & 11.13 & 6.73 & \textbf{2.61} \\
All \textit{E}/\textit{Z} bonds $\leq10^\circ$ $\uparrow$ & \underline{61.9} & 57.1 & \textbf{76.2} & \underline{59.6} & 59.2 & 59.5 & 42.8 & 58.1 & \textbf{79.1} \\
All \textit{R}/\textit{S} centers correct $\uparrow$ & \underline{73.4} & \textbf{100.0} & \textbf{100.0} & 51.7 & \underline{85.3} & \underline{85.3} & \textbf{85.7} & 85.1 & 85.1 \\
Success@1+stereo $\uparrow$ & 59.5 & \underline{81.1} & \textbf{82.7} & 25.4 & \underline{36.5} & \underline{36.5} & 24.0 & 36.1 & \textbf{45.2} \\
\bottomrule
\end{tabular}}
\vspace{-4pt}
\end{table}
\ifdefined\atomsteerpreprint
\begin{figure}[!b]
\else
\begin{figure}[!htb]
\fi
    \centering
    \includegraphics[width=0.80\textwidth,trim=4 3 7 3,clip]{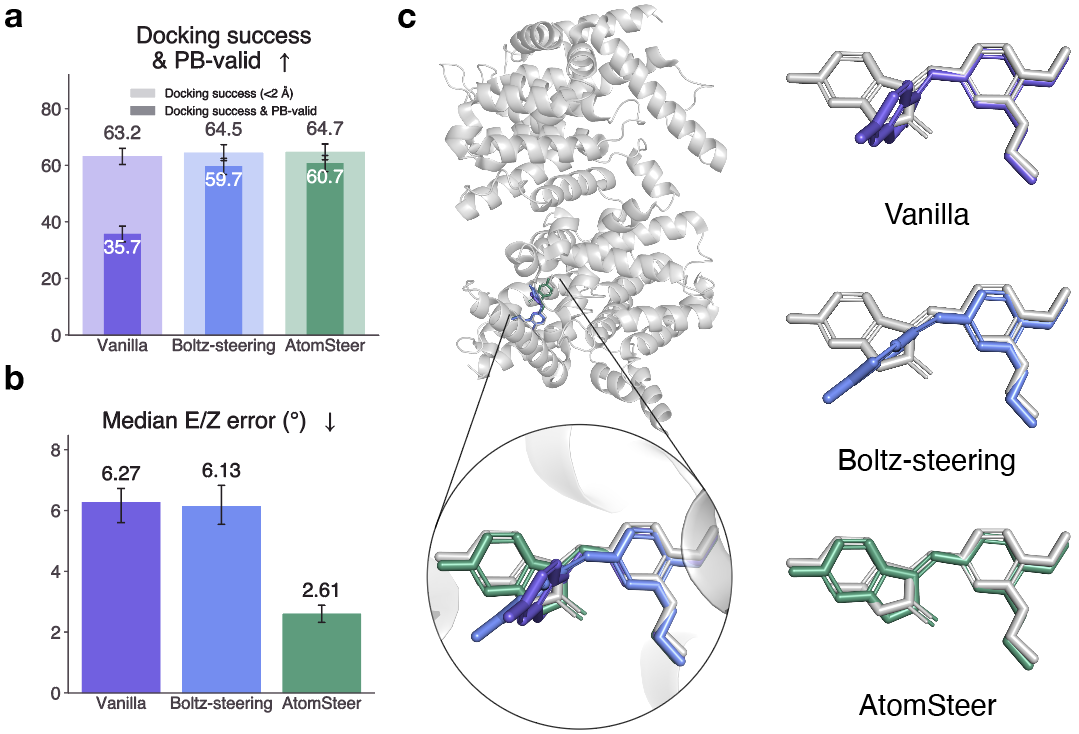}
    \caption{\textbf{Ligand stereochemistry and docking success on StereoBench.}
\textbf{(a)}~Docking success (ligand RMSD $<2$~\AA{} after pocket alignment, light)
and joint PB-valid docking success (Success@1, dark).
\textbf{(b)}~Median per-ligand \textit{E}/\textit{Z} dihedral error to the experimental ligand.
\textbf{(c)}~Example complex (PDB~\texttt{8ZQ2}): each method's pose overlaid on the
experimental ligand (gray), with an inset on the stereogenic bond.}
    \label{fig:boltz-stereobench}

\end{figure}

\looseness=-1
\textbf{Existing methods repair validity and chirality, but not double bond geometry.}
On StereoBench, every baseline substantially raises PoseBusters validity and \textit{R}/\textit{S} accuracy, yet none raises the within-$10^\circ$ fraction above the vanilla Boltz-2's $59.6\%$, and the approach from Ishitani \& Moriwaki, the most accurate on \textit{R}/\textit{S}, lowers the \textit{E}/\textit{Z} metric to $42.8\%$ (Table~\ref{tab:boltz-final}).

\looseness=-1
\textbf{AtomSteer corrects \textit{E}/\textit{Z} geometry.}
On StereoBench, AtomSteer raises the within-$10^\circ$ fraction to $79.1\%$, $+19.7$ points over Boltz-steering, and more than halves the median error, from $6.13^\circ$ to $2.61^\circ$ (Table~\ref{tab:boltz-final}; Figure~\ref{fig:boltz-stereobench}b). The same correction holds on the \textit{E}/\textit{Z} subset of the PoseBusters benchmark. In Figure~\ref{fig:boltz-stereobench}c, these three methods assign the correct descriptor but predict inaccurate dihedrals compared to the experimental geometry. Appendix Figures~\ref{fig:boltz_asbetter}, \ref{fig:boltz_asequal}, and~\ref{fig:boltz_asworse} show further corrections, unchanged cases, and also cases where AtomSteer displaces the ligand without a stereochemical gain.

\looseness=-1
\textbf{AtomSteer attains the best validity and docking success, and the highest joint success.}
On both benchmarks, AtomSteer has the highest PoseBusters validity, docking success, Success@1 (confidence top-1), and Success@5 (by ligand RMSD) of all evaluated methods (Table~\ref{tab:boltz-final}).
On StereoBench, PoseBusters validity rises from $51.0\%$ in vanilla Boltz-2 to $93.4\%$, Success@5 improves, and \textit{R}/\textit{S} accuracy is at parity with Boltz-steering (Table~\ref{tab:boltz-final}). Success@1+stereo, which requires the confidence top-1 pose to be PB-valid, docked below 2~\AA{}, within $10^\circ$ at every stereogenic double bond, and correct at every stereocenter, rises from $36.5\%$ under Boltz-steering to $45.2\%$. Protein structure quality is mostly indistinguishable across methods (Table~\ref{tab:stereobench-ost}).

\looseness=-1
\textbf{The \textit{E}/\textit{Z} gain holds after the training cutoff.}
On the 177 StereoBench complexes released after the Boltz-2 training cutoff, docking success falls for every method, whereas the \textit{E}/\textit{Z} improvement over Boltz-steering stays at $+19.2$ points (Table~\ref{tab:stereobench-holdout}), consistent with the correction enforcing the stereochemistry specified in the input graph rather than recalling complexes seen during training. On the PoseBusters benchmark, whose entries all pre-date the training cutoff, both steering methods reach $99.3\%$ validity and $100\%$ \textit{R}/\textit{S} accuracy, and AtomSteer raises the within-$10^\circ$ fraction from $57.1\%$ to $76.2\%$ on the 21 ligands with an eligible double bond. AtomSteer therefore enforces physical validity, and correct \textit{R}/\textit{S} and \textit{E}/\textit{Z} stereochemistry, while preserving the pose and folded structure produced by the pre-trained model.

\section{Conclusion}
\label{sec:conclusion}

\vspace{-5pt}\looseness=-1
AtomSteer provides a common test-time interface for structural constraints in
pre-trained atomistic generative models. Each requirement pairs structural residuals with structure-dependent projections. AtomSteer applies bounded corrections to the model's endpoint prediction and returns the corrected prediction through the native denoising sampler, so new requirements add constraint functions rather than requiring retraining the model. With this single interface, AtomSteer improves crystallographic symmetry and motif realization, conformer stereochemistry and quantum-chemical ensemble properties, as well as protein--ligand \textit{E}/\textit{Z} geometry, validity, and joint success over their pre-trained models. AtomSteer currently requires access to a predicted endpoint, or its useful estimate, during generation. Each requirement must be expressible as a structural residual with a corresponding projection. Extending the framework to broader diffusion and flow parameterizations, and to learned projections for less structured requirements, are natural next steps. AtomSteer could also accommodate further constraints, from coordination environments and oxidation states to experimentally derived structural information, enabling increasingly controllable generation across chemistry and materials design.

\subsection*{AI use statement}
In this work, we used generative AI tools for writing and editing code, running experiments, drafting parts of the paper, and plotting results. We have not used generative AI tools for helping to develop theoretical models and conceptual frameworks, or to propose and refine hypotheses, or for the design of or feedback on research methodology and experiments, or cleaning and reformatting datasets, or qualitative or thematic analysis. Generation of synthetic data sets, translation, formulation of mathematical claims, and providing ingredients for or writing proofs are not applicable to this work. We have reviewed all AI-assisted work. We take responsibility for the final content of this work, including text, claims or artifacts produced with the aid of generative AI.

\subsection*{Reproducibility statement}
We have put a lot of effort into ensuring that our results are reproducible. The AtomSteer framework, its constraint residuals, and its projection families are defined in Section~\ref{sec:method} and Appendix~\ref{app:shared-projections}. Their integration with each pre-trained backbone, including endpoint prediction, native update, projection order, and schedules, is summarized in Table~\ref{tab:atomsteer-backbone-schedules} and Algorithm~\ref{alg:atomsteer-sampling} and specified in Appendix~\ref{app:crystal-methods} for FlowMM, Appendix~\ref{app:conformer-methods} for ET-Flow, and Appendix~\ref{app:protein-ligand-methods} for Boltz-2. All experiments use frozen pre-trained backbone checkpoints and the released checkpoints of the baselines, with the versions, checkpoints, and sampling settings stated in Appendices~\ref{app:crystalline-materials}, \ref{app:molecular-conformers}, and~\ref{app:protein-ligand}, with checkpoint identifiers and source revisions to be provided with the code. The evaluation metrics are defined together with their thresholds, tolerances, and evaluation subsets: Appendix~\ref{app:crystal-metrics} covers crystalline materials, Appendix~\ref{app:conformer-metrics} covers molecular conformers, and Appendix~\ref{app:protein-ligand-metrics} covers protein--ligand complexes. The construction and filtering criteria of the StereoBench benchmark are described in Appendix~\ref{app:protein-ligand-tasks}. Together, the appendices specify the formulation, projection families, schedules, and evaluation protocols, and the code and configuration files will contain every numerical setting. We will release the codebase on GitHub upon publication.

\subsection*{Acknowledgments}
We acknowledge the MIT SuperCloud and Lincoln Laboratory Supercomputing Center for providing HPC resources. P.C. is supported by the Regenerative Energy-Efficient Manufacturing of Thermoset Polymeric Materials (REMAT) Energy Frontier Research Center, funded by the U.S. Department of Energy, Office of Science, Basic Energy Sciences under award DE-SC0023457. K.J. would like to acknowledge the funding support from the Defense Advanced Research Projects Agency (DARPA, BTO), (Agreement HR00112590006). B.D. would like to acknowledge the funding support from Shell Inc. N.S. would like to acknowledge the funding support from the Eli and Dorothy Berman Fellowship.
This material is based upon work supported by the U.S. National Science Foundation under award Nos CNS-2346520,  RISE-2425761, and DMS-2325184, by the Defense Advanced Research Projects Agency (DARPA) under Agreement No. HR00112490488,  by the Department of Energy, National Nuclear Security Administration under Award Number DE-NA0004266 and by the United States Air Force Research Laboratory under Cooperative Agreement Number FA8750-19-2-1000.  Neither the United States Government nor any agency thereof, nor any of their employees, makes any warranty, express or implied, or assumes any legal liability or responsibility for the accuracy, completeness, or usefulness of any information, apparatus, product, or process disclosed, or represents that its use would not infringe privately owned rights. Reference herein to any specific commercial product, process, or service by trade name, trademark, manufacturer, or otherwise does not necessarily constitute or imply its endorsement, recommendation, or favoring by the United States Government or any agency thereof. The views and opinions of authors expressed herein do not necessarily state or reflect those of the United States Government or any agency thereof.

\bibliography{references}

\bibliographystyle{atomsteer_style}

\newpage
\appendix
\raggedbottom
\makeatletter
\setlength{\@fptop}{0pt}
\setlength{\@fpsep}{16pt plus 2pt}
\setlength{\@fpbot}{0pt plus 1fil}
\makeatother
\FloatBarrier

\section{AtomSteer sampling and projection details}
\label{app:shared-details}

We first describe the shared sampling loop and molecular projections in AtomSteer.
Appendices~\ref{app:crystalline-materials}, \ref{app:molecular-conformers},
and~\ref{app:protein-ligand} give more details on the 3 domains discussed in this paper: crystalline materials, molecular conformers, and protein--ligand complexes.

\subsection{Sampling loop and backbone schedules}
\label{app:sampling}

AtomSteer uses the sampling loop in
Algorithm~\ref{alg:atomsteer-sampling}. At each step, one model evaluation
gives the endpoint prediction. Scheduled projections apply bounded corrections (with acceptance checks), and the native sampler update then continues from the corrected structure. Table~\ref{tab:atomsteer-backbone-schedules} lists, for each task, the endpoint prediction, the active projection families, and when they act along the
sampling trajectory. AtomSteer retains each pre-trained backbone's released sampler:
ET-Flow uses 50 flow matching steps, FlowMM 1{,}000 steps, and Boltz-2 200
diffusion steps. Appendices~\ref{app:crystal-methods}, \ref{app:conformer-methods},
and~\ref{app:protein-ligand-methods} specify the FlowMM, ET-Flow, and Boltz-2
implementations, respectively. 

\begin{algorithm}[H]
\caption{AtomSteer sampling from noise to structure (one trajectory)}
\label{alg:atomsteer-sampling}
\label{alg:etflow}
\label{alg:boltz-atomsteer}
\begin{algorithmic}[1]
\REQUIRE Pre-trained model; sampling times $(t_0,\ldots,t_K)$; constraint schedules (Table~\ref{tab:atomsteer-backbone-schedules}).
\STATE $x_{t_0}\sim\rho_{\rm init}$ \qquad (initial noise from the backbone's prior).
\FOR{$k=0,\ldots,K-1$}
  \STATE $\hat x_{t_k}^{\rm end}\gets\hat x_\theta^{\rm end}(x_{t_k},t_k)$ \qquad (predict endpoint).
  \STATE $\tilde x_{t_k}^{\rm end}\gets(\Pi_{c_m}\circ\cdots\circ\Pi_{c_1})(\hat x_{t_k}^{\rm end})$ \qquad (project endpoint).
  \STATE $x_{t_{k+1}}\gets
  \operatorname{Step}_{t_k\to t_{k+1}}
  (x_{t_k},\tilde x_{t_k}^{\rm end};x_{t_0})$.
\ENDFOR
\FOR{each terminal projection $c$, in projection order}
  \STATE $x_{t_K}\gets\Pi_c(x_{t_K})$.
\ENDFOR
\STATE \textbf{return} $x_{t_K}$.
\end{algorithmic}
\smallskip
{\footnotesize
In step 3, the endpoint prediction at $t=t_k$ is
\[
 \hat x_t^{\rm end}=\begin{cases}
 x_t+(1-t)v_\theta(x_t,t),&\text{ET-Flow},\\
 \operatorname{projx}[x_t+(1-t)v_\theta(x_t,t)],&\text{FlowMM},\\
 D_\theta(x_\sigma,\sigma),&\text{Boltz-2}.
 \end{cases}
\]
Boltz-2 uses the noisy coordinates $x_\sigma$ and noise level $\sigma$ passed to its denoiser. The scheduled constraints $c_1,\ldots,c_m$ act in order, and with no constraints projection applied at a particular step, $\tilde x_{t_k}^{\rm end}=\hat x_{t_k}^{\rm end}$.
$\operatorname{Step}$ reuses the model evaluation and follows
Table~\ref{tab:atomsteer-backbone-schedules}. Each $\Pi_c$ includes movement bounds and acceptance checks. An unchanged prediction or rejected projection keeps the native sampler's update.\par}
\end{algorithm}

\begin{table}[H]
\centering
\caption{\textbf{AtomSteer instantiations on the three backbones.}
Every task runs Algorithm~\ref{alg:atomsteer-sampling}. Percentages give the
share of sampling steps completed when a projection acts (FlowMM 1{,}000,
ET-Flow 50, and Boltz-2 200 steps). Projections act at every step within
each range unless a cadence is specified. Ranges ending at 100\% include
the last sampling update.
``Final'' means after the last sampling update. The schedule column lists molecular projections in their per-step order. Note that a projection happens only when a constraint is applicable to the structure. In FlowMM, the native sampler's velocity scaling is adopted so AtomSteer applies on top of what FlowMM intended: $Q(A,F,\ell)=(1,1+5t,1+5t)$ for de novo and motif-constrained generation and
$(1,1+10t,1)$ for both CSP variants.}
\label{tab:atomsteer-backbone-schedules}
\label{tab:flowmm-method-schedules}
\label{tab:boltz-schedule}
\small
\setlength{\tabcolsep}{3pt}
\begin{tabularx}{\textwidth}{>{\raggedright\arraybackslash}p{.12\textwidth}>{\raggedright\arraybackslash}p{.14\textwidth}>{\raggedright\arraybackslash}p{.20\textwidth}>{\raggedright\arraybackslash}X>{\raggedright\arraybackslash}p{.14\textwidth}}
\toprule
Task & Endpoint prediction & Active projection families & Projection schedule & Continuation \\
\midrule
FlowMM de novo & Velocity extrapolation & Crystal symmetry: inferred orbits and lattice metric & 80--100\%: every 2.5\% & Flow continuation \eqref{eq:atomsteer-flowmm-continuation} \\
\addlinespace[5pt]
FlowMM CSP & Velocity extrapolation & Crystal symmetry: lattice-metric template (inferred or supplied) and inferred orbits & \textit{Composition-only:}\newline Lattice and coordinates: 90\%, 100\%\newline \textit{Lattice-informed:}\newline Lattice: every 1\%\newline Coordinates: 90\%, 100\% & Flow continuation \eqref{eq:atomsteer-flowmm-continuation} \\
\addlinespace[5pt]
FlowMM motif & Velocity extrapolation & Crystal motif: species, positions, and in-plane cell & Every step & Initial-noise interpolation (Appendix~\ref{app:flowmm-continuation}) \\
\addlinespace[5pt]
ET-Flow conformers & Velocity extrapolation & ChemGN; rigid-fragment \textit{E}/\textit{Z}; global parity selection & \textit{E}/\textit{Z} fragment flips: 16--80\%\newline \textit{E}/\textit{Z} fragment refinement: 32--100\%\newline Cartesian ChemGN: 16--100\%\newline Final: ChemGN, then parity selection & Initial noise interpolation \eqref{eq:atomsteer-flow-reinsertion} \\
\addlinespace[5pt]
Boltz-2 complexes & Denoised EDM estimate & ChemGN (Cartesian and bond-tangent \textit{R}/\textit{S}); rigid-fragment \textit{E}/\textit{Z}; rigid-body ligand and protein components & \textit{E}/\textit{Z} fragment flips: 15--80\%\newline \textit{E}/\textit{Z} fragment refinement: 32.5--100\%\newline Cartesian ChemGN: 15--100\%\newline Ligand $SE(3)$: 67.5--100\%\newline Protein-component $SE(3)$: 60--100\%\newline Bond-tangent ChemGN (\textit{R}/\textit{S}): every step\newline Final: ligand, protein components, then \textit{R}/\textit{S} & EDM update \eqref{eq:atomsteer-edm-reinsertion} \\
\bottomrule
\end{tabularx}
\end{table}

\subsection{Molecular constraints and projections}
\label{app:shared-projections}

We derive geometric constraints from the input molecular graph and enforce
its specified stereochemistry. Residuals measure violations of these
requirements at the endpoint prediction, and the projections reduce
those residuals. ET-Flow and Boltz-2 use these shared constraints, with
their implementation details in
Appendices~\ref{app:conformer-methods} and~\ref{app:protein-ligand-methods}.

\subsubsection{Bond geometry, steric separation, and planarity}
\label{app:molecular-geometry-residuals}

\paragraph{Bond lengths.}
A covalent bond $(i,j)$ must have a length within the graph-derived interval
$[\bar\ell_{ij},\bar u_{ij}]$, determined by its atom and bond types.
With Cartesian atom positions $x_i,x_j$ and bond length
$d_{ij}=\|x_i-x_j\|_2$, its residual is
\begin{equation}
 r_{ij}^{\rm distance}(x)=
 \begin{pmatrix}
 [\bar\ell_{ij}-d_{ij}]_+\\ {}[d_{ij}-\bar u_{ij}]_+
 \end{pmatrix},\qquad [z]_+=\max(z,0).
 \label{eq:atomsteer-distance-residual}
\end{equation}
The components measure how much the bond is shorter than its lower bound
or longer than its upper bound.

\paragraph{Bond angles.}
The angle $\theta_{ijk}$ between bonds $(i,j)$ and $(j,k)$ is constrained
through the 1--3 distance $d_{ik}$ between atoms separated by two bonds:
\begin{equation}
 d_{ik}^2=d_{ij}^2+d_{jk}^2-2d_{ij}d_{jk}\cos\theta_{ijk}.
 \label{eq:atomsteer-angle-distance}
\end{equation}
We apply Eq.~\eqref{eq:atomsteer-distance-residual} to the graph-derived
distance bounds for these pairs. Reducing these 1--3 distance violations
together with the bond-length residuals constrains the bond angles.

\paragraph{Intramolecular steric clashes.}
Nonbonded atoms should remain sufficiently separated. For each eligible pair
$(i,j)$, the graph-derived minimum separation $\bar\ell_{ij}^{\rm nb}$ gives
\begin{equation}
 r_{ij}^{\rm clash}(x)=[\bar\ell_{ij}^{\rm nb}-d_{ij}]_+.
 \label{eq:atomsteer-clash-residual}
\end{equation}
The residual measures how much the pair distance $d_{ij}$ falls below this
minimum. Note that bond lengths and bond angles use their own distance constraints, while protein--ligand and intercomponent exclusions use atom-type radii and the
surrounding protein (Appendix~\ref{app:complex-constraints}).

\paragraph{Ring and double bond planarity.}
Aromatic rings and the atoms around a double bond should be approximately
coplanar. For each designated planar group $S$, its current coordinates
determine the centroid $\bar x_S$ and unit normal $n_S$ of the plane
minimizing squared perpendicular atom distances. With allowed distance
$\varepsilon_S$ from that plane, we use
\begin{equation}
 \begin{aligned}
 \bar x_S&=\frac1{|S|}\sum_{i\in S}x_i,
 &n_S&\in\arg\min_{\|n\|_2=1}\sum_{i\in S}[(x_i-\bar x_S)^\top n]^2,\\
 r_i^{\rm plane}&=[|(x_i-\bar x_S)^\top n_S|-\varepsilon_S]_+.
 \end{aligned}
 \label{eq:atomsteer-plane-residual}
\end{equation}
Here $|(x_i-\bar x_S)^\top n_S|$ is atom $i$'s perpendicular distance.
The residual measures the amount above $\varepsilon_S$. Ring and double
bond groups are specified separately.

\subsubsection{Double bond stereochemistry and tetrahedral chirality}
\label{app:stereo-residuals}

\paragraph{\textit{E}/\textit{Z} configuration.}
A stereogenic double bond $j=k$ must match the \textit{E}/\textit{Z}
configuration specified in the input (typically prescribed in a SMILES string). The ordered tuple $(i,j,k,l)$ uses substituents $i,l$ at double-bond atoms $j,k$. Let $u$ be the unit double-bond axis and $v,w$ the perpendicular components of the substituent
bond vectors. Writing
$u=(x_k-x_j)/\|x_k-x_j\|_2$,
$v=(x_i-x_j)-[(x_i-x_j)^\top u]u$, and
$w=(x_l-x_k)-[(x_l-x_k)^\top u]u$, we use
\begin{equation}
 \phi_{ijkl}=\operatorname{atan2}[(u\times v)^\top w,v^\top w],\qquad
 e_{ijkl}=\operatorname{atan2}[\sin(\phi_{ijkl}-\phi_\star),
                              \cos(\phi_{ijkl}-\phi_\star)].
 \label{eq:atomsteer-stereo-angle}
\end{equation}
Here $\phi_{ijkl}$ is the signed dihedral. The input determines the atom
ordering, with target $\phi_\star=\pi$ for \textit{E} and $\phi_\star=0$
for \textit{Z}. The wrapped error $e_{ijkl}$ measures the angular deviation
and treats $-\pi$ and $\pi$ identically. We reduce
$r^{\rm EZ}=e_{ijkl}/s_\phi$, which measures this error in units of the
angular scale $s_\phi$.

\paragraph{Tetrahedral handedness.}
Each tetrahedral stereocenter must match the input graph's
\textit{R}/\textit{S} configuration (typically prescribed in a SMILES string). For four graph-ordered atom positions, the oriented volume is proportional to
\begin{equation}
 V_{ijkl}=(x_i-x_l)^\top[(x_j-x_l)\times(x_k-x_l)].
 \label{eq:atomsteer-oriented-volume}
\end{equation}
Reflection reverses its sign, and the graph ordering identifies the
requested handedness. For local correction, let $\phi_i$ be stereocenter
$i$'s signed improper dihedral and $s_i\in\{-1,1\}$ its graph-derived
orientation sign. With signed margin $m_i=s_i\phi_i$ and required positive
angular margin $\mu$, we use
\begin{equation}
 r_i^{\rm RS}(x)=[\mu-m_i(x)]_+
 \label{eq:atomsteer-rs-residual}
\end{equation}
This residual measures the violation of the required signed-angle margin.
In ET-Flow, graph-ordered four-substituent dihedral margins select the
conformer's global parity after sampling
(Eq.~\eqref{eq:atomsteer-parity-projection}).

\subsubsection{Structure-dependent molecular projections}
\label{app:molecular-projection-solvers}

Implementations on molecular projections use two recurring projection families.
ChemGN reduces local chemical residuals through Cartesian or bond-tangent
updates. Rigid-fragment and rigid-body projections move connected fragments
or complete components while preserving their internal geometry. Global parity
selection, a final step that chooses between a conformer
and its mirror image, is a separate operation.

\paragraph{Cartesian ChemGN.}
The Cartesian form of ChemGN reduces active local chemical residuals by
correcting selected Cartesian coordinates.
Let $q$ contain those coordinates and $J=\partial r/\partial q$ be the
residual Jacobian. A damped linearized correction is
\begin{equation}
 \Delta q=\arg\min_d\frac12\|r+Jd\|_2^2+\frac\epsilon2\|d\|_2^2
 =-J^\top(JJ^\top+\epsilon I)^{-1}r.
 \label{eq:atomsteer-chemgn}
\end{equation}
Residual normalization is specified in Appendices~\ref{app:conformer-methods}
and~\ref{app:complex-constraints}. We apply the update separately to each
residual family over a bounded set of active residuals. A blend factor
$\beta$ and movement setting $\Delta_{\max}$ bound the resulting correction.
This is the local residual-reduction step underlying the molecular
instantiations of Eq.~\eqref{eq:atomsteer-relaxed-projection}.

\paragraph{Bond-tangent ChemGN for local \textit{R}/\textit{S}.}
Within a protein complex, we correct ligand chirality while protecting its
bond geometry. This is the ChemGN update of Eq.~\eqref{eq:atomsteer-chemgn} with a
tangent-restricted Jacobian. Let $B$ be the Jacobian of the protected bond distances. We
restrict the chirality Jacobian using
\begin{equation}
 T_{\eta_{\rm tan}}=I-B^\top(BB^\top+\eta_{\rm tan}I)^{-1}B,
 \qquad \bar J=J_{\rm RS}T_{\eta_{\rm tan}}.
 \label{eq:atomsteer-bond-tangent}
\end{equation}
The Gauss--Newton solve uses $\bar J$ to reduce
Eq.~\eqref{eq:atomsteer-rs-residual}. This regularized tangent restriction
limits first-order bond changes. An update is accepted only if it keeps
already-correct \textit{E}/\textit{Z} assignments and \textit{R}/\textit{S} signs and stays within the displacement
bounds.

\paragraph{\textit{E}/\textit{Z} fragment flips and refinement.}
Correcting a double bond orientation can require moving an entire substituent
rather than only the four atoms used to measure its dihedral. Let $\mathcal W$
be an eligible connected fragment, with bond axis $u$ through $p$. We apply
\begin{equation}
 x_i'=\begin{cases}
 p+R_u(\vartheta)(x_i-p),&i\in\mathcal W,\\
 x_i,&i\notin\mathcal W.
 \end{cases}
 \label{eq:atomsteer-ez-projection}
\end{equation}
The rotation preserves every distance within $\mathcal W$. Graph connectivity,
ring closures, and covalent attachments determine which fragments can rotate.
\textit{E}/\textit{Z} fragment flips move the fragment into the requested stereochemical
basin. \textit{E}/\textit{Z} fragment refinement then uses ADAM for bounded rotations
of the same fragment. \textit{E}/\textit{Z} ChemGN reduces the remaining
dihedral error through local coordinate corrections.

\paragraph{Rigid-body component projection.}
A ligand or protein component can reduce overlap with its environment through
$x_i'=Rx_i+b$, with $(R,b)\in SE(3)$. Proper rigid motion preserves the
component's internal distances and stereochemistry. We optimize its translation
and rotation against the intercomponent residuals in
Appendix~\ref{app:complex-constraints}. Covalently connected chains move
together, preserving their internal connections.

\paragraph{Global conformer parity.}
For an isolated conformer, we compare its existing orientation with a
reflection through a plane passing through its centroid:
\begin{equation}
 x_i'=\bar x+H(x_i-\bar x),\qquad H=\operatorname{diag}(1,1,-1).
 \label{eq:atomsteer-parity-projection}
\end{equation}
This changes global handedness while preserving pairwise distances, bond
angles, and planarity. We then choose the orientation with more accurate signed stereocenters using the graph-oriented dihedral margins. This projection chooses global parity rather than independently changing individual stereocenters.

\clearpage

\subsection{Comparison with Related Methods}
\label{app:related_methods}

Table~\ref{tab:method_comparison} compares the design choices underlying representative
inference-time steering and structure-aware generation methods.

\DeclareRobustCommand{\capyes}{{\usefont{U}{pzd}{m}{n}\char51}}
\DeclareRobustCommand{\capno}{\textcolor{black!45}{\usefont{U}{pzd}{m}{n}\char55}}
\DeclareRobustCommand{\cappartial}{\tikz[baseline=-0.55ex]{\draw[line width=0.55pt] (0,0) circle (0.75ex); \fill (0,0) -- (90:0.75ex) arc (90:270:0.75ex) -- cycle;}}
\begin{table}[!h]
\centering
\caption{\textbf{Design choices in constrained generative modeling.}
Comparison of representative approaches to inference-time control
and structure-aware generation.
\capyes{} denotes a demonstrated capability,
\cappartial{} a partial or variant-dependent capability,
and \capno{} a capability not demonstrated in the reported method.}
\label{tab:method_comparison}

\small
\setlength{\tabcolsep}{4pt}
\renewcommand{\arraystretch}{1.13}

\resizebox{\linewidth}{!}{%
\begin{tabular}{lcccc}
\toprule
\textbf{Method}
& \shortstack{\textbf{No}\\\textbf{retraining}}
& \shortstack{\textbf{Test-time}\\\textbf{steering}}
& \shortstack{\textbf{Structure-aware}\\\textbf{correction}}
& \shortstack{\textbf{Multiple atomistic}\\\textbf{domains}} \\
\midrule

Feynman--Kac steering~\citep{singhal2025fk}
    & \capyes & \capyes & \capno & \capno \\

Boltz-steering~\citep{passaro2025boltz2}
    & \capyes & \capyes & \cappartial & \capno \\

Restraint-guided Boltz~\citep{ishitani2025stereo}
    & \capyes & \capyes & \cappartial & \capno \\

PCFM~\citep{utkarsh2025physics}
    & \capyes & \capyes & \capno & \capno \\

Projected Diffusion~\citep{christopher2024projected}
    & \capyes & \capyes & \capno & \capno \\

Constrained protein diffusion~
\citep{christopher2026constraineddiffusionproteindesign}
    & \capyes & \capyes & \cappartial & \capno \\

\citet{chen2026gs}
    & \cappartial & \cappartial & \capyes & \capno \\

SCIGEN~\citep{okabe2026scigen}
    & \capyes & \capyes & \cappartial & \capno \\

DiffCSP++~\citep{jiao2024diffcsppp}
    & \capno & \capno & \cappartial & \capno \\

SymmCD~\citep{levy2025symmcd}
    & \capno & \capno & \cappartial & \capno \\

\midrule
\textbf{AtomSteer (ours)}
    & \capyes & \capyes
    & \capyes & \capyes \\
\bottomrule
\end{tabular}%
}
\end{table}

\section{Crystalline materials}
\label{app:crystalline-materials}

We describe the FlowMM implementation, evaluation protocols, and additional
results for de novo generation, crystal structure prediction, and
motif-constrained generation.

\subsection{FlowMM integration and crystal projections}
\label{app:crystal-methods}
\paragraph{Endpoint prediction.}
We use FlowMM as a pre-trained backbone for symmetry and motif constraints in
crystal generation~\citep{miller2024flowmm}. We follow FlowMM's original training implementations to first pre-train a vector field for both de novo and CSP tasks on the MP-20 dataset, and use the de novo checkpoint for de novo and motif-constrained generation tasks and CSP checkpoint for the CSP related tasks. The model learns a Riemannian vector field over atom types, fractional positions, and
lattice parameters, with sampling state $x_t=(A_t,F_t,\ell_t)$ and
$F_t\in\mathbb T^{N\times3}$. Atom types use continuous analog bits in de novo
and motif-constrained generation, and composition is prescribed in CSP. FlowMM represents the lattice by $\ell$, comprising three cell lengths and
transformed cell angles. The physical cell $L$ contains the lattice vectors as
rows and has metric $G=LL^\top$. Projections use $L$ and $G$, and sampling
updates use $\ell$. 
All tasks follow FlowMM's anti-annealing settings for each task and sampling with 1{,}000 Euler steps. At time $t$ and next time $\tau$, one model evaluation gives the raw field, the native
update, and the endpoint prediction,
\begin{equation}
 \begin{aligned}
 v_{\rm raw}&=v_\theta(x_t,t),&
 x_\tau^{\rm nat}&=\operatorname{projx}
 [x_t+(\tau-t)Q(t)v_{\rm raw}],\\
 \hat x_t^{\rm end}&=\operatorname{projx}
 [x_t+(1-t)v_{\rm raw}]&&\text{endpoint extrapolation}.
 \end{aligned}
 \label{eq:atomsteer-flowmm-endpoint}
\end{equation}
where $Q(t)$ is FlowMM's velocity scaling following their paper's implementations
(Table~\ref{tab:flowmm-method-schedules}). 
For fractional coordinates,
$\Delta_{\mathbb T}(F,H)=(H-F)-\operatorname{round}(H-F)$ measures
periodic displacement in each component, and $\wrap(u)=u-\lfloor u\rfloor$
maps each component into $[0,1)$.

\subsubsection{Crystal symmetry projections}
\label{app:periodic-residuals}

Crystal symmetry projection couples two requirements: symmetry-related
fractional coordinates and a compatible lattice metric.

\paragraph{Fractional-coordinate symmetry.}
Crystal symmetry must map atoms onto atoms of the same species modulo unit-cell
translations. For a realization with a fixed cell setting, let $R_g,b_g$
be the fractional rotation and translation of operation $g$, and $\pi_g(i)$
the same-species partner assigned to atom $i$. For fractional column
positions $f_i$, we use
\begin{equation}
 r_{F,g,i}=\Delta_{\mathbb T}(f_{\pi_g(i)},R_gf_i+b_g).
 \label{eq:atomsteer-orbit-residual}
\end{equation}
The residual measures the periodic positional mismatch between the
transformed atom and its assigned partner.

\paragraph{Lattice-metric symmetry.}
Lattice symmetry restricts the unit-cell lengths and angles, with seven
lattice systems listed in Table~\ref{tab:metric-constraints}. The cell must
make each selected fractional rotation $R_g$ preserve physical distances.
For a cell $L$ with lengths $a,b,c$ and angles $\alpha,\beta,\gamma$, the
metric and residual are
\begin{equation}
 G=LL^\top=
 \begin{pmatrix}
 a^2&ab\cos\gamma&ac\cos\beta\\
 ab\cos\gamma&b^2&bc\cos\alpha\\
 ac\cos\beta&bc\cos\alpha&c^2
 \end{pmatrix},\qquad r_{G,g}=R_g^\top G R_g-G.
 \label{eq:atomsteer-periodic-constraint}
\end{equation}
Here $R_g$ is the rotation matrix of symmetry operation $g$ in fractional
coordinates, and $r_{G,g}$ measures its change to the cell metric.
We correct the atomic positions and cell geometry to support the same
selected symmetry.

Hierarchy-aware symmetry selection infers compatible realizations from the
endpoint prediction. Symmetry detection over a range of tolerances and cell
settings enumerates the realizations that the endpoint prediction
approximately satisfies. Selection starts from the feasible realization that
needs the least projection work, follows minimal-supergroup relations to the
most symmetric feasible parent, and uses projection work only to choose
between parents that are not related by inclusion.

Within an atomic orbit, the predicted images determine a fractional
representative $q_0$. After fixing integer translations, operations that
leave it on the same periodic site give affine constraints $Bq=b$.
We project the representative and reconstruct its symmetry-related sites by
\begin{equation}
 q^\star=q_0+B^\dagger(b-Bq_0),\qquad
 f_{\pi_g(i)}^\star=R_gq^\star+b_g\pmod{\mathbb Z^3}.
 \label{eq:atomsteer-orbit-reconstruction}
\end{equation}
Here $B,b$ collect the positional requirements, $B^\dagger$ is the
pseudoinverse, and $q^\star$ is the corrected representative.

For de novo generation, let $\mathcal S_a$ be the selected realization's
rotation set and $\hat G=\hat L\hat L^\top$ the predicted metric:
\begin{equation}
 G^{\rm av}=\frac1{|\mathcal S_a|}\sum_{R\in\mathcal S_a}R^\top\hat G R,
 \qquad
 G^\star=\left(\frac{\det\hat G}{\det G^{\rm av}}\right)^{1/3}G^{\rm av}.
 \label{eq:atomsteer-metric-projection}
\end{equation}
The average $G^{\rm av}$ satisfies the selected metric symmetry, and the
determinant rescaling would preserve the flow model's predicted cell volume.
A Procrustes lift constructs a cell with metric $G^\star$ near the predicted
orientation. Atom types and order remain fixed. If no eligible realization
is found, the original sampling update is used.

\paragraph{Lattice-metric constraints in CSP.}
CSP constrains cell lengths and angles to the selected lattice system and
axis convention in Table~\ref{tab:metric-constraints}. We express the
predicted cell $\hat L$ in the conventional setting and form its metric
$G_c$. For metric template $\mathcal T$, the projection is
\begin{equation}
 G_c^\star=P_{\mathcal T}(G_c).
 \label{eq:atomsteer-csp-metric}
\end{equation}
Here $P_{\mathcal T}$ projects the independent metric entries onto the
required relations. Factoring the positive-definite metric $G_c^\star$
gives cell vectors in the conventional setting. We transform them back to
the original setting and uniformly rescale them to preserve the volume
of $\hat L$.

\begin{table}[t]
\centering
\caption{\textbf{Lattice-metric constraints.}
Residual components measure deviations from the metric relations of each family in the stated setting.}
\label{tab:metric-constraints}
\small
\begin{tabularx}{\textwidth}{l>{\raggedright\arraybackslash}X}
\toprule
Metric family & Entries of the constraint residual \\
\midrule
Triclinic & No additional metric equalities \\
Monoclinic ($b$ unique) & $G_{12},\ G_{23}$ \\
Orthorhombic & $G_{12},\ G_{13},\ G_{23}$ \\
Tetragonal ($c$ unique) & $G_{12},\ G_{13},\ G_{23},\ G_{11}-G_{22}$ \\
Cubic & $G_{12},\ G_{13},\ G_{23},\ G_{11}-G_{22},\ G_{22}-G_{33}$ \\
Hexagonal ($c$ unique) & $G_{13},\ G_{23},\ G_{11}-G_{22},\ 2G_{12}+G_{11}$ \\
Rhombohedral & $G_{11}-G_{22},\ G_{22}-G_{33},\ G_{12}-G_{13},\ G_{13}-G_{23}$ \\
\bottomrule
\end{tabularx}
\end{table}

Here $G_{ij}$ is the dot product of lattice vectors $i,j$. Off-diagonal
entries constrain angles, and diagonal differences constrain relative
lengths. The projector $P_{\mathcal T}$ projects the independent metric
entries onto these relations by least squares, setting required entries to
zero and fitting the common parameter of related entries.

\paragraph{Composition-only and lattice-informed CSP.}
In composition-only CSP, hierarchy-aware symmetry selection infers the
metric template, its setting, and the orbit realization from the endpoint
prediction, then corrects the lattice and the fractional coordinates.

Lattice-informed CSP receives a reference-derived metric template, which fixes
one of the seven lattice systems in Table~\ref{tab:metric-constraints} and its
axis convention, and an integer setting transform to the conventional cell
(FlowMM is trained on Niggli-reduced cells). FlowMM generates the numerical lengths, angles, volume, and fractional
coordinates. The lattice is projected onto this template, and hierarchy-aware
symmetry selection infers coordinate symmetry from the lattice-corrected
endpoint prediction and projects only the fractional coordinates. An
incompatible coordinate projection is skipped.

\subsubsection{Motif-constrained projections}
\label{app:motif-residuals}

We apply motif projections to SCIGEN's Archimedean lattice (AL)
task~\citep{okabe2026scigen}.
Each condition specifies a motif, atomic species, and nearest-neighbor distance
$d$. The constraint fixes the designated atoms' species, their periodic
arrangement, and the cell's in-plane dimensions. The pre-trained generative model generates the remaining atoms and cell parameters.

\paragraph{Specified atomic species.}
For motif-constrained generation, the designated sites $i\in S$ must
contain the requested element with atomic number $Z$. FlowMM stores each atom's identity as a continuous vector $A_i$.
Let $a_Z$ be its vector representation of the requested element.
The residual $r_{A,i}=A_i-a_Z$ measures disagreement with that element. Projection then assigns $A_i=a_Z$ at the motif sites.

\paragraph{Periodic atomic pattern.}
For motif-constrained generation, designated atoms must form the requested
triangular, honeycomb, or Kagome pattern in a common lattice plane. Their basal fractional sites are
$\{(0,0)\}$ for triangular,
$\{(1/3,2/3),(2/3,1/3)\}$ for honeycomb, and
$\{(0,0),(1/2,0),(0,1/2)\}$ for Kagome.
Let $m_i$ be these sites in a common fractional plane and $\zeta$ their
shared translation. For predicted positions $f_i$, we use
\begin{equation}
 r_{F,i}^{\rm AL}(F,\zeta)=\Delta_{\mathbb T}(f_i,m_i+\zeta),\qquad
 \zeta^\star\in\arg\min_{\zeta\in\mathbb T^3}
 \sum_{i\in S}\|r_{F,i}^{\rm AL}(F,\zeta)\|_2^2.
 \label{eq:atomsteer-motif-positions}
\end{equation}
The residual measures each atom's periodic displacement from its
translated motif site. The least-squares translation $\zeta^\star$
places the pattern near the predicted atoms, giving
$f_i^\star=\wrap(m_i+\zeta^\star)$.

\paragraph{Absolute lattice scale.}
The cell must realize the requested nearest-neighbor separation $d$ with
basal lengths $a=b=sd$ and angle $\gamma=2\pi/3$. The geometric factor
$s$ is $1,\sqrt3,2$ for triangular, honeycomb, and Kagome patterns,
respectively. We use
\begin{equation}
 r_L^{\rm AL}(L)=
 (a-sd,\ b-sd,\ \gamma-2\pi/3).
 \label{eq:atomsteer-motif-cell}
\end{equation}
The components measure two basal-length deviations and one angular
deviation. Projection then sets $a=b=sd$ and $\gamma=120^\circ$, retaining
the remaining cell parameters $c,\alpha,\beta$.

\subsubsection{Projection schedule and acceptance checks}
\label{app:crystal-validity}
Atoms and their periodic images must remain sufficiently separated and contact checks use the physical cell $L$. In de novo generation and CSP,
each trial cell is uniformly rescaled to the uncorrected cell's volume
at the same sampling stage, then converted back to FlowMM's lattice
parameters. For fractional positions $f_i,f_j$, the shortest separation
and its residual are
\begin{equation}
 d_{ij}^{\rm PBC}=\min_{n\in\mathbb Z^3}\|(f_j-f_i-n)L\|_2,
 \qquad r_{ij}^{\rm contact}=[d_0-d_{ij}^{\rm PBC}]_+,
 \label{eq:atomsteer-periodic-contact}
\end{equation}
Here $n$ denotes integer lattice translations and $d_0$ the minimum
separation. The residual measures how far the periodic distance falls
below $d_0$ and is used to check corrections.
Every trial cell must be finite, exceed a minimum volume, and remain well
conditioned, and in de novo generation and CSP its conditioning must stay
close to that of the native cell. The lattice parameters must also reconstruct the proposed physical cell
within numerical tolerance. A failed check rejects the correction, and the affected coordinates keep the native sampler's update. 

\paragraph{Numerical settings.}
The numerical bounds in Table~\ref{tab:atomsteer-flowmm-settings} limit atomic motion and
cell deformation during correction.
Endpoint bounds apply to the predicted clean structure. CSP also checks
the correction after it enters the next sampling state.
\begin{table}[H]
\centering
\caption{Numerical settings for AtomSteer projections in FlowMM.}
\label{tab:atomsteer-flowmm-settings}
\small
\setlength{\tabcolsep}{4pt}
\renewcommand{\arraystretch}{1.08}
\begin{tabularx}{\textwidth}{@{}>{\raggedright\arraybackslash}p{0.30\textwidth}>{\raggedright\arraybackslash}X@{}}
\toprule
Setting & Value \\
\midrule
Symmetry detection &
Distance tolerances $0.05,0.075,0.10,0.15,0.20,0.30$~\AA{};
angle tolerance $5^\circ$ \\
Scales for comparing corrections &
Atomic displacement scale $0.20$~\AA{}; cell deformation scale $0.10$ (logarithmic strain) \\
\addlinespace[3pt]
Symmetry-candidate search &
Maximum atomic displacement $0.10$~\AA{} in CSP \\
Endpoint correction &
Maximum atomic displacement $0.20$~\AA{} in de novo generation and CSP \\
CSP next-state correction &
Maximum atomic displacement $0.10$~\AA{};
maximum absolute logarithmic stretch $0.10$ \\
\addlinespace[3pt]
Periodic contacts &
$d_0=0.5$~\AA{}, with the CSP acceptance rule above \\
\bottomrule
\end{tabularx}
\end{table}

\subsubsection{Flow continuation}
\label{app:flowmm-continuation}
FlowMM continues sampling from the projected endpoint. In de novo
generation and CSP, the next state is
\begin{equation}
 x_\tau=x_\tau^{\rm nat}+\tau\,(\tilde x_t^{\rm end}-\hat x_t^{\rm end}),
 \label{eq:atomsteer-flowmm-continuation}
\end{equation}
where the fractional coordinates use the periodic displacement
$\Delta_{\mathbb T}$ and are wrapped; entries that the projection does not
change keep the native update. Motif-constrained generation
continues along an interpolation with the initial noise $x_0$ and the projected endpoint.
\FloatBarrier
\subsection{Datasets and evaluation protocols}
\label{app:crystal-metrics}
\label{app:crystal-tasks}

\subsubsection{MP-20 de novo generation}
\label{app:crystal-denovo-protocol}

For each of 10{,}000 paired samples, in our experiments, vanilla FlowMM and AtomSteer use the
same initial noise and atom count.

\paragraph{Symmetry metrics.}
We detect space groups in as-generated and UMA-relaxed structures with
spglib~\citep{togo2024spglib} at $\mathrm{symprec}=0.10$~\AA{} and $5^\circ$
angle tolerance.
We define the \textit{symmetry success rate} as the fraction of the full paired
cohort assigned a space-group number of at least three, excluding $P1$ and
$P\bar{1}$, so a structure without a detected space group is counted as a failure. The distribution distance $d_{\mathrm{SG}}$ is the base-2 Jensen--Shannon
distance between the detected generated and MP-20 space-group distributions.
Only structures with a detected space group enter this comparison and the
unique-SG count, and lower $d_{\mathrm{SG}}$ is better.

\paragraph{Geometric validity.}
An as-generated structure is geometrically valid (Valid) when its fractional
coordinates and cell are finite, the cell is nonsingular, and its minimum
periodic interatomic distance, including an atom's own periodic images, is at
least 0.5~\AA{}, the distance cutoff used in FlowMM~\citep{miller2024flowmm}. We evaluate all 10{,}000 samples before relaxation and count unevaluable
structures as invalid.

\paragraph{Forces and full-cell relaxation.}
Forces and relaxations use the UMA-S-1p2 potential (OMat
task)~\citep{wood2025uma}. For a structure $x$ with UMA atomic forces
$\mathbf F_i(x)$, $F_{\max}(x)=\max_i\|\mathbf F_i(x)\|_2$. The paired statistic
$\Delta F_{\max}$ is the median over the 10{,}000 pairs of
$F_{\max}(x^{\mathrm{AtomSteer}})-F_{\max}(x^{\mathrm{FlowMM}})$, computed on
as-generated structures before relaxation, with FlowMM as the reference. Full-cell relaxation starts from every as-generated structure and optimizes atomic positions and all cell parameters with the LBFGS optimizer, a Fr\'echet cell filter, force threshold 0.05~eV/\AA{}, and a 100-step limit,
without external pressure or symmetry constraints.

\subsubsection{MP-20 crystal structure prediction}
\label{app:crystal-csp-protocol}

The MP-20 testset contains 9{,}046 targets, and each method we evaluate generates 20 samples (for $K=20$ evaluations). DiffCSP, CrystalFlow, and OMatG use their released code and MP-20 CSP checkpoints.
For DiffCSP, its $1{,}000$-step diffusion sampler with
\texttt{step\_lr}=$10^{-5}$ is used, CrystalFlow uses 100 Euler steps with coordinate
anti-annealing slope 5, and OMatG uses its Linear-ODE checkpoint and 210 Euler
steps. FlowMM and both AtomSteer variants start from the same saved initial noise for each target.

\paragraph{Structure matching and space-group recovery.}
Structure match metrics adopt the validity-aware StructureMatcher
protocol of DiffCSP~\citep{jiao2023diffcsp} and FlowMM~\citep{miller2024flowmm}.
A candidate matches its reference only when both pass the FlowMM validity
check and the matcher returns a finite RMS distance. The validity check
requires a minimum periodic interatomic distance of at least 0.5~\AA{}, a cell
volume of at least 0.1~\AA$^3$, SMACT charge neutrality with the Pauling
electronegativity test (single elements and all-metal compositions pass), and
a successful CrystalNN local-environment fingerprint (crystal construction and
preprocessing follow the released FlowMM evaluator). The matcher uses
\texttt{stol=0.5}, \texttt{ltol=0.3}, \texttt{angle\_tol=10}. 

For a target $i$ with generated candidates $g_{ij}$,
\begin{align}
\mathrm{MR@}K
&=\frac{1}{N}\sum_i \mathbf{1}\!\left[\exists j<K:\,
    \mathrm{match}(g_{ij},r_i)\right],\\
\mathrm{SG@}K
&=\frac{1}{N}\sum_i \mathbf{1}\!\left[\exists j<K:\,
    \mathrm{SG}(g_{ij})=\mathrm{SG}(r_i)\right],\\
\mathrm{Joint@}K
&=\frac{1}{N}\sum_i \mathbf{1}\!\left[\exists j<K:\,
    \mathrm{match}(g_{ij},r_i)\land
    \mathrm{SG}(g_{ij})=\mathrm{SG}(r_i)\right].
\end{align}
Space groups are detected independently for candidate and reference structure at
$\mathrm{symprec}=0.10$~\AA{} and $5^\circ$ angle tolerance. Joint@K requires at least one candidate to both match the reference structure and recover its space group. For each target, @1 evaluates the first generated candidate, while @20 counts success if any of the 20 candidates meets the criterion.

\subsubsection{Motif-constrained generation}
\label{app:crystal-motif-protocol}

For each of Kagome, honeycomb, and triangular Archimedean lattices (ALs),
we draw 1{,}000 conditions following SCIGEN's condition-sampling procedure
and released AL-specific atom-count and bond-length distributions.
The element is drawn uniformly from Mn, Fe, Co, Ni, Ru, Nd, Gd, Tb, Dy,
and Yb. The atom count satisfies $k<N\leq20$, where $k$ is the number of
designated AL sites. Vanilla FlowMM and AtomSteer also start from the same initial state for each condition, while SCIGEN shares the design conditions and runs its released code and MP-20
checkpoint with its unmodified 1{,}000-step diffusion sampler and
\texttt{step\_lr}=$5\times10^{-6}$. 

\paragraph{Full-cell relaxation.}
We evaluate the methods with a full-cell UMA relaxation: UMA-S-1p2 with its OMat
task, Fr\'echet cell filter, force threshold 0.05~eV/\AA{}, and
500-step limit, relaxing atomic positions and the full cell without external
pressure. 

\paragraph{AL constraint satisfaction.}
The AL constraint is satisfied when the requested magnetic element forms the
requested AL at the prescribed nearest-neighbor distance in a geometrically
valid crystal~\citep{okabe2026scigen}. A crystal is geometrically valid when it keeps
the requested atom count, has atomic numbers from 1 to 100, a cell volume of at
least 0.1~\AA$^3$, and a minimum periodic interatomic distance, including
self-images, of at least 0.5~\AA{}. The detector searches same-species subsets
under the 48 signed axis permutations, fits a common periodic translation, and
accepts a Cartesian positional RMS of at most 0.15~\AA{}, relative differences
of at most 0.03 between the in-plane lengths $a$ and $b$ and between each
length and the requested side length $sd$
(Appendix~\ref{app:motif-residuals}), and an in-plane angle within $2^\circ$ of
$120^\circ$.

\paragraph{Prescreening and joint success.}
We apply SCIGEN's released four-stage prescreening filter to every method:
SMACT charge neutrality, space occupancy ratio below 1.7, and the two
released GNN classifiers for energy above hull and for distinguishing
pristine from perturbed structures. Prescreening survival is the fraction of samples that pass all four stages, while joint success is the fraction that both satisfy the AL constraint and pass prescreening.

\paragraph{AL detection after relaxation.}
We report the fraction of conditions for which UMA relaxation converges to a geometrically valid structure containing the requested AL shape. All metrics are computed over the full 1{,}000 planned conditions per AL and method.

\FloatBarrier
\subsection{Additional results for crystalline materials generation}
\label{app:crystal-additional-results}

\noindent\textbf{De novo generation.}
AtomSteer lowers the maximum atomic force in 74.5\% of the 10{,}000
paired samples. Full-cell relaxation converges for 82.16\% of
AtomSteer structures versus 80.48\% for vanilla FlowMM.

\subsubsection{Crystal structure prediction by size}

\begin{figure}[!htb]
    \centering
    \includegraphics[width=0.6\textwidth]{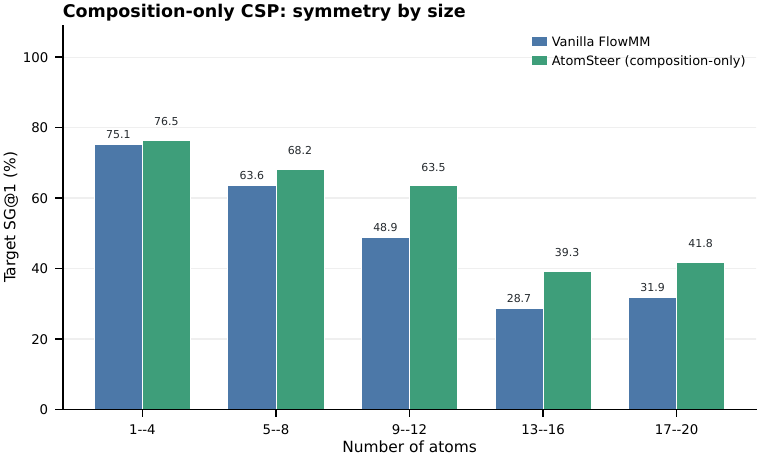}
    \caption{\textbf{Target-space-group recovery of composition-only AtomSteer
    by crystal size.} Target SG@1 is evaluated on all 9{,}046 MP-20 targets
    and grouped by the number of atoms in the generated unit cell.}
    \label{fig:flowmm-csp-by-size}
\end{figure}

\newpage
\subsubsection{Motif-constrained generation examples}

\begin{figure}[!htb]
    \centering
    \includegraphics[width=0.74\textwidth]{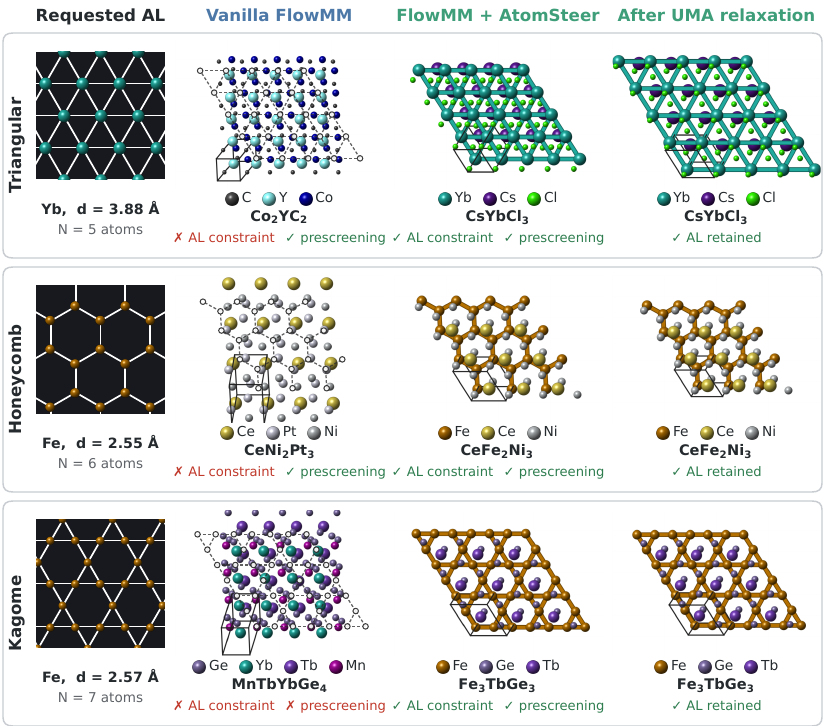}
    \caption{\textbf{Motif-constrained generation examples.} Each row fixes
    the target element, nearest-neighbor distance $d$, and atom count $N$, then
    shows the requested AL, matched vanilla FlowMM and AtomSteer structures, and
    the AtomSteer structure after UMA relaxation. Structures use a shared tilt
    and length scale within each row.}
    \label{fig:flowmm-motif-examples}
\end{figure}

\clearpage
\section{Molecular conformers}
\label{app:molecular-conformers}

We describe the ET-Flow implementation, GEOM-Drugs evaluation protocols,
and additional conformer results.

\subsection{ET-Flow integration and molecular projections}
\label{app:conformer-methods}
\paragraph{Endpoint prediction.}
We use the released GEOM-Drugs checkpoint \texttt{drugs-o3} of
ET-Flow for molecular conformer generation~\citep{hassan2024etflow}.
The ET-Flow and ET-Flow-SS baselines use the released deterministic and
stochastic 50-step samplers, respectively, both with the harmonic prior
and post hoc chirality correction. ET-Flow-SS additionally uses churn
and noise scales of one for $0.8\le t<1$.
AtomSteer uses 50 Euler steps and replaces the post hoc chirality
correction with global parity selection.
Starting from initial noise $x_0\sim\rho_0$, the predicted velocity at
time $t$ defines the endpoint estimate and native Euler update to time $\tau$:
\begin{equation}
 \hat x_t^{\rm end}=x_t+(1-t)v_\theta(x_t,t),\qquad
 x_\tau^{\rm nat}=x_t+(\tau-t)v_\theta(x_t,t).
 \label{eq:atomsteer-flow-endpoint}
\end{equation}

\paragraph{Molecular constraints.}
We use the molecular geometric residuals in Appendix~\ref{app:molecular-geometry-residuals}, normalizing distance and clash violations by their padded bounds and planarity violations
by the allowed distance from the plane. For stereochemistry, Cartesian ChemGN reduces the \textit{E}/\textit{Z} dihedral residual $r^{\rm EZ}=e/s_\phi$ (Eq.~\eqref{eq:atomsteer-stereo-angle}).
\textit{E}/\textit{Z} fragment flips and refinement follow Appendix~\ref{app:molecular-projection-solvers}.

\paragraph{Projection schedule.}
Table~\ref{tab:atomsteer-backbone-schedules} specifies the projection
schedules. At each applicable step, \textit{E}/\textit{Z} fragment flips
and refinement precede Cartesian ChemGN, which then acts sequentially on
\textit{E}/\textit{Z}, bond lengths, 1--3 distances, intramolecular
clashes, and planarity. After the final flow update, terminal
Cartesian ChemGN is followed by global parity selection
(Eq.~\eqref{eq:atomsteer-parity-projection}). 

\paragraph{Bounded corrections and acceptance.}
\textit{E}/\textit{Z} fragment flips must reach the target angular band, while fragment refinement must reduce the dihedral error without substantially worsening other constraints. Terminal ChemGN checks \textit{E}/\textit{Z}, bond lengths, bond angles, clashes, and planarity. Table~\ref{tab:atomsteer-etflow-settings} gives the numerical settings and acceptance tolerances.

\begin{table}[H]
\centering
\caption{\textbf{Numerical settings for AtomSteer projections in ET-Flow.}
$\Delta x$ is the bounded ChemGN correction over all atom coordinates, and $\tau$ is the
next sampling time.}
\label{tab:atomsteer-etflow-settings}
\small
\setlength{\tabcolsep}{4pt}
\renewcommand{\arraystretch}{1.08}
\begin{tabularx}{\textwidth}{@{}>{\raggedright\arraybackslash}p{0.26\textwidth}>{\raggedright\arraybackslash}X@{}}
\toprule
Setting & Value \\
\midrule
Relative distance padding & Bonds and 1--3 pairs: 0.125. Intramolecular
clashes: 0.10 \\
Planarity & Allowed distance from the plane: 0.25~\AA{} for ring and double
bond groups. Cartesian residuals are excess distances divided by this tolerance \\
\textit{E}/\textit{Z} parameters & Angular scale $s_\phi=20^\circ$,
Cartesian residual $r^{\rm EZ}=e/s_\phi$, flip acceptance band $5^\circ$ \\
\addlinespace[3pt]
\textit{E}/\textit{Z} fragment refinement & Eight ADAM steps, learning rate
0.05, maximum applied fraction 0.75. \newline
Rotation $\le30^\circ$ and RMS displacement $\le0.75$~\AA{} over moved
fragment atoms. \newline
Acceptance: increase $\le0.01$ in each checked constraint type's maximum
absolute residual and residual $L_2$ norm, including \textit{R}/\textit{S} \\
\addlinespace[3pt]
ChemGN & One step and up to 64 largest absolute residuals per
stage, damping $10^{-6}$. \newline
During sampling, bound $\|\Delta x\|_2\le(0.45/\tau)$~\AA{}, then apply
30\% of the bounded correction \\
\addlinespace[3pt]
ChemGN (terminal) & Combined coordinate norm $\|\Delta x\|_2\le0.04$~\AA{}. \newline
Acceptance: increase $\le10^{-6}$ in each checked constraint type's
maximum squared residual \\
\bottomrule
\end{tabularx}
\end{table}

\paragraph{Native continuation.}
When an endpoint correction is accepted, AtomSteer interpolates between
the initial noise $x_0$ and the projected endpoint. Otherwise,
sampling follows the native Euler update:
\begin{equation}
 x_\tau=\begin{cases}
 (1-\tau)x_0+\tau\tilde x_t^{\rm end},
       &\tilde x_t^{\rm end}\ne\hat x_t^{\rm end},\\
 x_\tau^{\rm nat},&\tilde x_t^{\rm end}=\hat x_t^{\rm end}.
 \end{cases}
 \label{eq:atomsteer-flow-reinsertion}
\end{equation}

\FloatBarrier
\subsection{Dataset and evaluation protocols}
\label{app:conformer-metrics}
\label{app:conformer-tasks}

\subsubsection{GEOM-Drugs test set and conformer budgets}
\label{app:conformer-data}

Following the released ET-Flow evaluator, we exclude two molecules
with disconnected graphs from the 1{,}000-molecule GEOM-Drugs test set.
The remaining 998 molecules contain 103{,}966 reference conformers,
with 207{,}932 generated conformers per method. All methods in
Table~\ref{tab:etflow-main} use these matched ensembles.

\subsubsection{Conformer accuracy, validity, and stereochemistry}
\label{app:conformer-geometric-metrics}

\paragraph{Coverage and matching.}
Following ET-Flow~\citep{hassan2024etflow}, we evaluate conformer
ensembles using coverage (COV) and average minimum RMSD (AMR).
For each molecule, we compute pairwise symmetry-corrected heavy-atom
RMSDs between reference and generated conformers. Recall coverage (COV-R) measures the fraction of reference conformers whose nearest generated conformer is within 0.75~\AA{}, while AMR-R averages these minimum RMSDs. Precision metrics (COV-P and AMR-P) reverse the comparison,
measuring how closely generated conformers match the reference ensemble. All metrics are computed per molecule and summarized by the mean and median across 998 molecules, weighting each molecule equally.
Table~\ref{tab:etflow-main} reports means, while Table~\ref{tab:etflow-cov-amr-full} reports both statistics.

\paragraph{Conformer validity.}
We evaluate validity using PoseBusters 0.6.3 with the \texttt{mol}
configuration~\citep{buttenschoen2024posebusters}. A conformer must
pass all twelve checks: molecule loading, sanitization, InChI
conversion, connectivity, absence of radicals, bond lengths,
bond angles, internal steric clashes, aromatic-ring flatness,
non-aromatic-ring non-flatness, double bond flatness, and internal
energy. Unevaluable checks count as failures. For each molecule
with $K$ reference conformers, we evaluate the first $\min(2K,32)$
generated conformers and report the mean validity rate across molecules.

\paragraph{Stereochemical distribution error.}
For each stereogenic double bond, we measure the unsigned angle
in $[0^\circ,180^\circ]$ between substituent bond vectors projected
perpendicular to the double bond, with $0^\circ$ and $180^\circ$
corresponding to planar cis and trans placements.
We compute the one-dimensional Wasserstein distance between generated
and held-out GEOM dihedral distributions, average over evaluable bonds
within each molecule, and report the median across 153 eligible molecules. A bond is evaluable when RDKit assigns an \textit{E}/\textit{Z}
configuration after hydrogen removal and each double-bond atom
has a heavy-atom substituent, following the same criterion as
Appendix~\ref{app:protein-ligand-metrics}.

\subsubsection{GFN2-xTB evaluation of ensemble properties}
\label{app:conformer-xtb-protocol}
Following ET-Flow~\citep{hassan2024etflow}, we evaluate four
GFN2-xTB ensemble properties: Boltzmann-weighted means of energy $E$,
dipole moment $\mu$, and HOMO--LUMO gap $\Delta\epsilon$, together with
minimum energy $E_{\min}$. We report the median absolute error across
molecules relative to GEOM reference ensembles, using
$kT=0.592$~kcal/mol for Boltzmann weighting.
We randomly select 100 molecules from the sorted GEOM-Drugs test
set using a fixed seed of 42. For each molecule with $K$ reference
conformers, we retain the first $\min(K,32)$ reference and
$\min(2K,32)$ generated conformers. Generated conformers are geometrically optimized using GFN2-xTB
6.7.1~\citep{bannwarth2019gfn2xtb} with normal convergence criteria,
followed by single-point calculations with a maximum of 1{,}000
electronic iterations and dipole evaluation. Reference conformers
undergo the same single-point calculations without optimization.
Molecular charges and unpaired-electron counts are specified
where applicable. All 100 molecules have complete evaluations
for every method.
\FloatBarrier
\subsection{Additional results}
\label{app:conformer-additional-results}

\subsubsection{Conformer coverage and matching statistics}

\begin{table}[H]
\centering
\caption{\textbf{Conformer coverage and matching across 998 GEOM-Drugs
test molecules.} Each metric is computed per molecule, with means and medians
reported across molecules. COV uses a 0.75~\AA{} threshold. Best results are
\textbf{bold} and second-best results are \underline{underlined}.}
\label{tab:etflow-cov-amr-full}
\scriptsize
\setlength{\tabcolsep}{4.0pt}
\begin{tabular}{@{}lrrrrrrrr@{}}
\toprule
& \multicolumn{2}{c}{COV-R (\%) $\uparrow$}
& \multicolumn{2}{c}{AMR-R (\AA{}) $\downarrow$}
& \multicolumn{2}{c}{COV-P (\%) $\uparrow$}
& \multicolumn{2}{c}{AMR-P (\AA{}) $\downarrow$} \\
\cmidrule(lr){2-3}\cmidrule(lr){4-5}\cmidrule(lr){6-7}\cmidrule(lr){8-9}
Method & Mean & Median & Mean & Median & Mean & Median & Mean & Median \\
\midrule
ET-Flow
& \underline{79.92} & \underline{84.52} & 0.458 & \underline{0.436}
& 73.40 & \underline{80.31} & 0.562 & 0.500 \\
ET-Flow-SS
& \textbf{79.93} & \textbf{84.57} & \textbf{0.456} & \textbf{0.435}
& \underline{73.52} & 80.19 & \underline{0.559} & \underline{0.497} \\
AtomSteer
& 79.01 & 82.90 & \underline{0.457} & 0.440
& \textbf{75.49} & \textbf{82.65} & \textbf{0.529} & \textbf{0.463} \\
\bottomrule
\end{tabular}
\end{table}

\subsubsection{Qualitative conformer results}

\begin{figure}[!htb]
    \centering
    \includegraphics[width=0.9\textwidth]{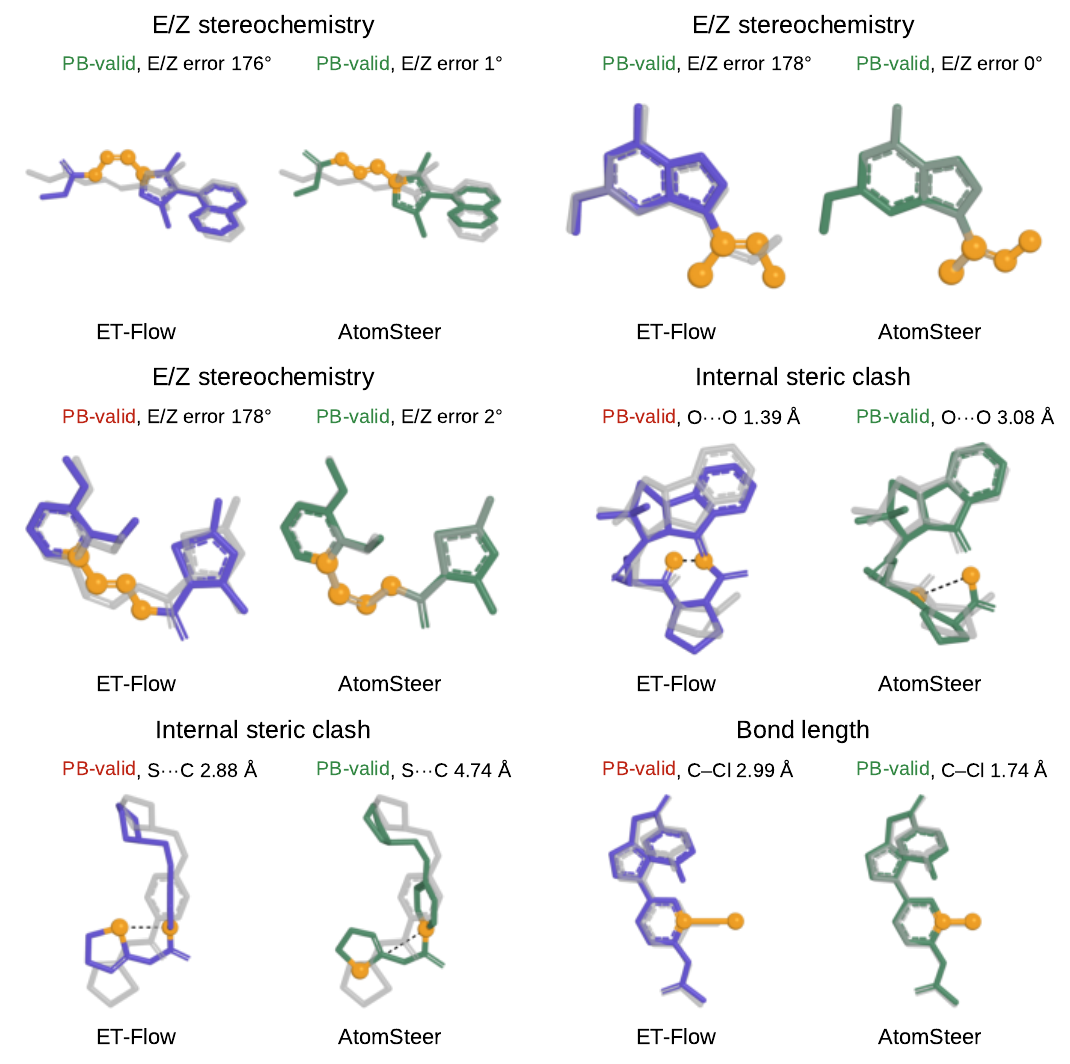}
    \caption{\textbf{Examples of constraint corrections on GEOM-Drugs.}
    ET-Flow (purple) and AtomSteer (green) conformers are aligned to a common
    GEOM reference (gray). Highlighted atoms indicate the evaluated constraint.
    Annotations report PoseBusters validity (green = valid, red = invalid) and
    the \textit{E}/\textit{Z} dihedral error, nonbonded distance, or bond length.
    PoseBusters settings here do not check \textit{E}/\textit{Z} configuration, so
    stereochemical violations may still pass its checks. Each example pairs an
    ET-Flow violation with a nearby PB-valid AtomSteer conformer satisfying the
    highlighted constraint.}
    \label{fig:etflow-fixes}
\end{figure}

\begin{figure}[!htb]
    \centering
    \includegraphics[width=\textwidth]{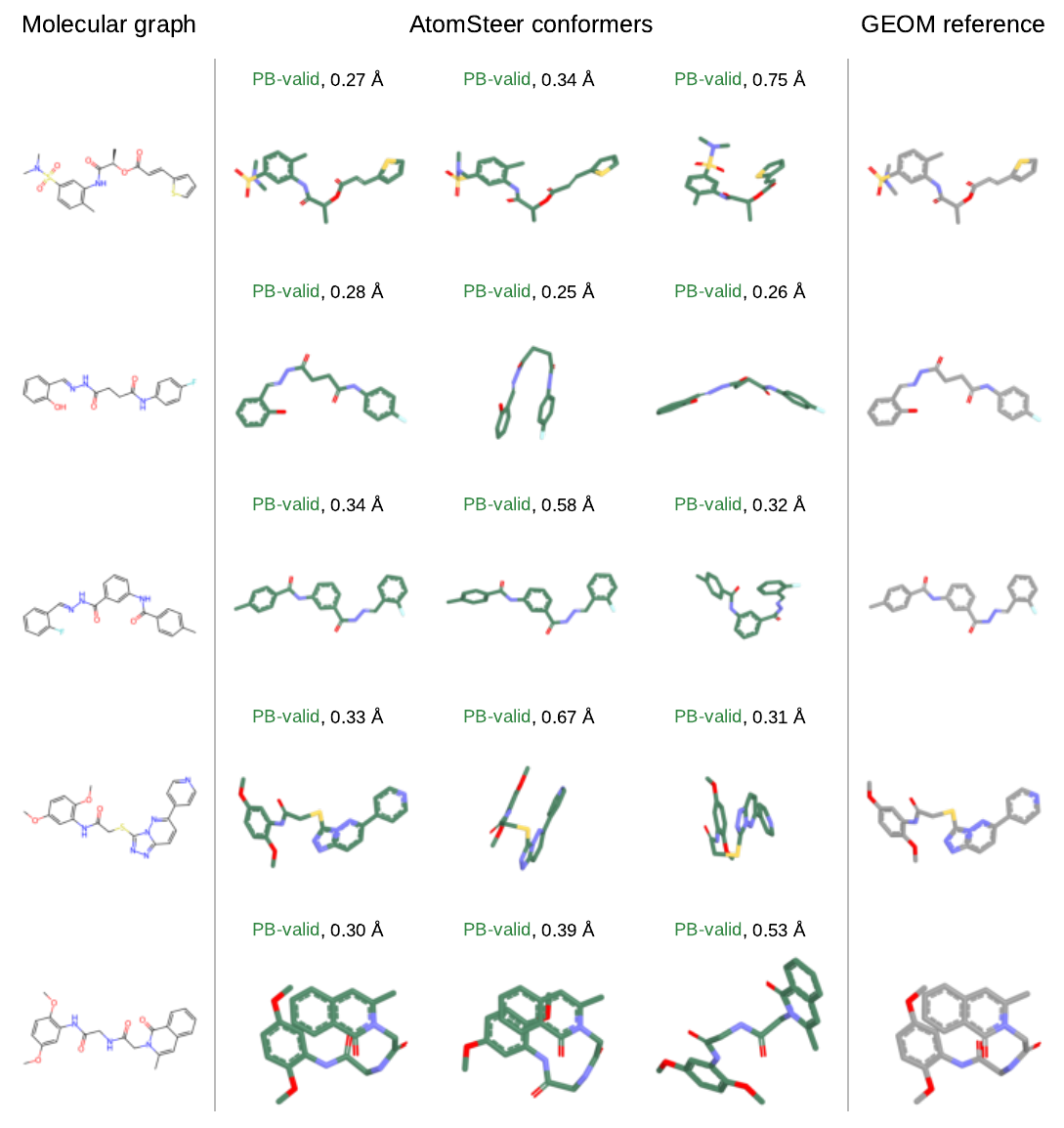}
    \caption{\textbf{AtomSteer conformer ensembles for five GEOM-Drugs
    molecules.} Each row shows the molecular graph, three distinct AtomSteer
    conformers, and the GEOM reference nearest to the first. Annotations report
    PoseBusters validity and symmetry-corrected heavy-atom RMSD to each
    conformer's nearest reference. All shown conformers are PB-valid, with
    nearest-reference RMSDs of 0.25--0.75~\AA{}.}
    \label{fig:etflow-conformers}
\end{figure}

\clearpage
\section{Protein--ligand complexes}
\label{app:protein-ligand}

\subsection{Boltz-2 integration and comparison methods}
\label{app:protein-ligand-methods}

\paragraph{Protein--ligand generation with Boltz-2.}
We use Boltz-2 as a pre-trained backbone for applying molecular and
intermolecular constraints to protein--ligand complexes~\citep{passaro2025boltz2}.
The denoiser predicts all-atom coordinates from a noisy complex. Atom-to-token
mappings, bonds, stereochemical labels, and component memberships specify the
relations used by AtomSteer. AtomSteer reads the \textit{E}/\textit{Z} and \textit{R}/\textit{S}
labels from the input isomeric SMILES, which is identical for all methods we evaluate. We retain Boltz-2 v2.2.1 with its default recycling and MSA settings, 200 denoising steps, FP32 arithmetic, seed 42, and five returned diffusion samples per target. The top-1 sample is selected by native Boltz confidence. The sampling step scale is $s_{\rm step}=1.5$. Each trajectory uses one particle and AtomSteer projections in
place of Boltz-steering's potential guidance. Beyond the common supplied
molecular graph, no method uses experimental coordinates or reference
stereochemical labels during generation or selection.
Boltz-2 follows EDM~\citep{karras2022edm}, with unscaled noisy coordinates
$x_\sigma=x_0+\sigma\epsilon$, $\epsilon\sim\mathcal N(0,I)$, and $\sigma$
decreasing to $0$ during sampling. Its denoised estimate
$\hat x_{0,k}^{\rm EDM}=D_\theta(x_{\hat\sigma_k},\hat\sigma_k)$ is the
endpoint prediction $\hat x_{\hat\sigma_k}^{\rm end}$.

\paragraph{Native continuation.}
Boltz-2 keeps its native schedule of $K=200$ noise levels decreasing to
$\sigma_K=0$, its stochastic churn, which raises the noise level from
$\sigma_k$ to $\hat\sigma_k\ge\sigma_k$ before each denoiser call. AtomSteer projects
$\hat x_{0,k}^{\rm EDM}$ to $\tilde x_{0,k}^{\rm EDM}$. Native reverse
alignment then maps the noisy coordinates to this prediction's frame,
$\bar x_{\hat\sigma_k}=\mathsf{Align}_{\rm native}
(x_{\hat\sigma_k},\tilde x_{0,k}^{\rm EDM})$, and the native EDM update with
step scale $s_{\rm step}$ continues from the projected endpoint:
\begin{equation}
 x_{\sigma_{k+1}}=\bar x_{\hat\sigma_k}
 +s_{\rm step}(\sigma_{k+1}-\hat\sigma_k)
 \frac{\bar x_{\hat\sigma_k}-\tilde x_{0,k}^{\rm EDM}}{\hat\sigma_k}.
 \label{eq:atomsteer-edm-reinsertion}
\end{equation}
If the endpoint is unchanged, this is the original Boltz-2 sampling update.

\subsubsection{Ligand and intermolecular constraints}
\label{app:complex-constraints}

\paragraph{Ligand bond lengths and angles.}
We use the molecular distance constraints of
Appendix~\ref{app:molecular-geometry-residuals}. The ligand's atom types,
bond orders, and connectivity define lower and upper distance bounds
$\ell_{ij},u_{ij}$ for bonded and 1--3 pairs. With relative padding $\eta$,
we use
\begin{equation}
 \begin{aligned}
 \bar\ell_{ij}&=(1-\eta)\ell_{ij},&
 \bar u_{ij}&=(1+\eta)u_{ij},\\
 r_{ij}^{\rm ligand}&=
 \left(\frac{[\bar\ell_{ij}-d_{ij}]_+}{\bar\ell_{ij}},
       \frac{[d_{ij}-\bar u_{ij}]_+}{\bar u_{ij}}\right).
 \end{aligned}
 \label{eq:atomsteer-ligand-normalization}
\end{equation}
Here $d_{ij}$ is the pair distance and $\bar\ell_{ij},\bar u_{ij}$ the
padded bounds. The components measure relative violations of these bounds.
Intramolecular clashes use the lower-bound component with their own
padding.

\paragraph{Ligand planarity and stereochemistry.}
Ligand ring and double bond planarity use
Eq.~\eqref{eq:atomsteer-plane-residual}, with the plane computed from the
atoms' current coordinates. The \textit{E}/\textit{Z} dihedral error is
defined in Eq.~\eqref{eq:atomsteer-stereo-angle}. For local ligand
chirality correction, bond-tangent ChemGN (\textit{R}/\textit{S}) reduces
the signed-margin residual in Eq.~\eqref{eq:atomsteer-rs-residual} with
the bond-tangent correction in Eq.~\eqref{eq:atomsteer-bond-tangent}.

\paragraph{Protein--ligand exclusion.}
Nonbonded ligand and protein atoms must remain sufficiently separated.
For ligand atom $i$ and protein atom $j$ with van der Waals radii
$\rho_i,\rho_j$ and relative padding $\eta_{\rm PL}$, we use
\begin{equation}
 d_{ij}^{\min}=(1-\eta_{\rm PL})(\rho_i+\rho_j),\qquad
 r_{ij}^{\rm PL}(x)=
 \frac{[d_{ij}^{\min}-\|x_i-x_j\|_2]_+}{d_{ij}^{\min}}.
 \label{eq:atomsteer-pl-exclusion}
\end{equation}
Here $d_{ij}^{\min}$ is the minimum separation. The residual measures the
amount by which the pair distance falls below it, normalized by
$d_{ij}^{\min}$. Eligible pairs exclude covalently connected and single-atom
chains and are limited to currently violated nearby pairs. Cartesian ChemGN
adjusts ligand coordinates while the protein is fixed. Rigid ligand
$SE(3)$ projection reduces the same residual by translating and rotating
the ligand as one body, preserving its internal geometry.

\paragraph{Protein-component separation.}
For complexes with independently moving protein components, we reduce
overlap between components. Covalently connected protein chains form one
rigid component. Atom pairs in different components use
Eq.~\eqref{eq:atomsteer-pl-exclusion} with the same padding. For components
$P,Q$ containing the same molecular entity, with centroids
$\bar x_P,\bar x_Q$, we also impose minimum centroid distance $d_{\rm cent}$:
\begin{equation}
 r_{PQ}^{\rm centroid}(x)=
 \frac{[d_{\rm cent}-\|\bar x_P-\bar x_Q\|_2]_+}{d_{\rm cent}}.
 \label{eq:atomsteer-component-centroid}
\end{equation}
The residual measures the relative violation of this centroid separation.
The ligand-contacting receptor component stays fixed. In complexes without covalent protein--nonprotein links, other eligible
components undergo bounded $SE(3)$ projections while preserving their
internal coordinates.

\subsubsection{Projection schedule and terminal projections}

\paragraph{Projection order and schedule.}
Each denoising step projects the endpoint prediction immediately before the
native update, in this order: \textit{E}/\textit{Z} fragment flips and refinement; Cartesian
ChemGN; the rigid-body ligand and protein-component $SE(3)$ projections; and
bond-tangent ChemGN (\textit{R}/\textit{S}). The fraction of the sampling trajectory in which each projection acts is given
in Table~\ref{tab:boltz-schedule}; the shared projections are defined in
Appendix~\ref{app:shared-projections}.

\paragraph{Bounded corrections and acceptance.}
Each projection is bounded. \textit{E}/\textit{Z} fragment refinement, rigid ligand $SE(3)$, and bond-tangent ChemGN (\textit{R}/\textit{S}) accept a correction only if it lowers their targeted residuals and passes checks on the other constraint families, and the protein-component projection limits any increase in protein--ligand exclusion. The settings in Table~\ref{tab:atomsteer-boltz-settings} control geometric tolerances and correction sizes,
with activation windows in Table~\ref{tab:atomsteer-backbone-schedules}.
For ChemGN, $\Delta x$ combines the endpoint correction across
all ligand coordinates, and
$\alpha_k=|s_{\rm step}(\sigma_{k+1}-\hat\sigma_k)/\hat\sigma_k|$
is its coefficient in Eq.~\eqref{eq:atomsteer-edm-reinsertion}.

\begin{table}[H]
\centering
\caption{\textbf{Numerical settings for AtomSteer projections in Boltz-2.}}
\label{tab:atomsteer-boltz-settings}
\small
\setlength{\tabcolsep}{4pt}
\renewcommand{\arraystretch}{1.08}
\begin{tabularx}{\textwidth}{@{}>{\raggedright\arraybackslash}p{0.26\textwidth}>{\raggedright\arraybackslash}X@{}}
\toprule
Setting & Value \\
\midrule
Relative distance padding & Bonds and 1--3 pairs: 0.125. Intramolecular
clashes: 0.10. Protein--ligand exclusion: 0.225 \\
Planarity & Allowed distance from the plane: 0.25~\AA{} for ring and double
bond groups, 0.02~\AA{} for terminal rings. Cartesian residuals are excess distances in \AA{} \\
\textit{E}/\textit{Z} parameters & Angular scale $s_\phi=20^\circ$,
Cartesian residual $r^{\rm EZ}=2e/s_\phi$, flip acceptance band $5^\circ$ \\
\addlinespace[3pt]
\textit{E}/\textit{Z} fragment refinement & Eight ADAM steps, learning rate
0.05, maximum applied fraction 0.75. \newline
Rotation $\le30^\circ$ and RMS displacement $\le0.75$~\AA{} over moved
fragment atoms. \newline
Acceptance: increase $\le0.01$ in each checked constraint type's maximum
absolute residual and residual $L_2$ norm, including \textit{R}/\textit{S} \\
\addlinespace[3pt]
Cartesian ChemGN & One step and up to 64 largest absolute residuals per
stage, damping $10^{-6}$. \newline
During sampling, bound $\|\Delta x\|_2\le(0.45/\alpha_k)$~\AA{}, then apply
30\% of the bounded correction \\
\addlinespace[3pt]
Rigid ligand & Translation $\le0.45$~\AA{}, rotation $\le5^\circ$,
RMS displacement $\le0.50$~\AA{} \\
Rigid protein component & Translation $\le0.40$~\AA{}, rotation
$\le1.5^\circ$, minimum same-entity centroid separation 1~\AA{} \\
Local \textit{R}/\textit{S} & Margin $30^\circ$, at most four inner updates. \newline
RMS displacement $\le0.10$~\AA{} per update, sum of RMS displacements
$\le0.75$~\AA{} per application of the local correction \\
\addlinespace[3pt]
ChemGN (terminal) & Combined coordinate norm $\le0.40$~\AA{} from the
ligand before the first terminal pass. \newline
Stopping: no constraint type's maximum squared residual decreases
by more than $10^{-4}$ \\
\bottomrule
\end{tabularx}
\end{table}

\paragraph{Terminal projection and confidence selection.}
After the last denoising step, terminal ligand ChemGN applies up to ten
passes with early stopping, followed by one bounded protein-component
correction and \textit{R}/\textit{S} correction. Confidence is then
predicted on the five final complexes, and the pose with the highest
confidence is selected as top-1.

\paragraph{Comparison methods.}
Vanilla Boltz-2 (unsteered) and Guidance only each use one particle, five returned
samples, and 200 steps. Guidance only applies the released Boltz-steering
potentials with 20 gradient steps per denoiser call. Boltz-steering uses
the same guidance with three interacting particles and Feynman--Kac
resampling~\citep{singhal2025fk,wohlwend2025boltz1,passaro2025boltz2},
with weight 4 and resampling every three steps.
The method from Ishitani \& Moriwaki uses the authors' Boltz-2 fork with up to 1,000
conjugate-gradient iterations per denoiser call and unit bond, angle,
chirality, and environment-exclusion weights~\citep{ishitani2025stereo}.
For \citet{chen2026gs}, we apply the released solver to
Boltz-2 outputs before confidence evaluation, with 100 terminal
Gauss--Seidel sweeps per sample, $\alpha=10^{-7}$, and weight 1.
All methods use the same checkpoint, seed, inputs, and native Boltz-2 confidence
selection.

\subsection{Benchmarks}
\label{app:protein-ligand-tasks}

\paragraph{StereoBench.}
StereoBench comprises 1{,}103 experimental protein--ligand complexes indexed by
PDB accession and ligand identifier \citep{berman2000pdb}, each containing at
least one non-ring stereogenic double bond. We apply quality filters
(resolution $\leq 2.5$~\AA{}, $R_\mathrm{free} \leq 0.30$, ligand RSCC $\geq 0.8$,
RSR $\leq 0.25$, occupancy $\geq 0.8$, MW 100 to 1000~Da, $\geq 6$ heavy atoms,
non-covalent, drug-like, and a complete crystallographic model). Reference
\textit{E}/\textit{Z} configurations are assigned from crystal geometry using
RDKit's CIP labeler; deposited coordinates are used only for post-generation
evaluation. \textit{E}/\textit{Z} metrics use all 1{,}103 targets, while
\textit{R}/\textit{S} metrics use the 538 targets with an evaluable stereocenter.

\paragraph{PoseBusters.}
Of 308 complexes, 3 exhaust GPU memory for every method and 4 lack an
identifiable crystallographic reference pocket, leaving 301 common targets.
Success@5 uses the 300 targets with complete PoseBusters and RMSD results for
all five poses under every method. The \textit{E}/\textit{Z} and
\textit{R}/\textit{S} subsets contain 21 and 177 targets, respectively. All
entries were released between 2021-01-13 and 2023-05-24, before the Boltz-2
training cutoff.

\subsection{Evaluation metrics}
\label{app:protein-ligand-metrics}

\paragraph{PoseBusters validity.}
We evaluate PoseBusters 0.6.5 in \texttt{redock} mode
\citep{buttenschoen2024posebusters}, checking intermolecular validity against
each method's predicted complex rather than the deposited receptor. PB-valid
requires passing the applicable checks in the 27-check profile, excluding
\texttt{rmsd\_le\_2A}, which compares unaligned predicted and reference ligands;
docking accuracy is evaluated separately after pocket alignment. The two
water checks pass vacuously for co-folding predictions without waters, as do
the four cofactor checks when no cofactor is predicted.

\paragraph{Docking and joint success.}
Docking RMSD is the symmetry-corrected ligand heavy-atom RMSD to the
experimental ligand after C$\alpha$ alignment of the predicted and reference
10~\AA{} binding pockets; docking success requires RMSD $<2$~\AA{}.
Top-1 selects the highest-confidence pose among five using the Boltz-2 score
$(4\,\mathrm{pLDDT}_{\mathrm{complex}}+\mathrm{ipTM})/5$.
For AtomSteer, confidence is evaluated on the five final complexes before
selection. Success@1 requires the top-1 pose to be PB-valid with RMSD
$<2$~\AA{}; Success@5 requires any of the five poses to meet both criteria and
is an oracle statistic. All other single-pose rows of
Table~\ref{tab:boltz-final} use top-1.

\paragraph{Stereochemistry.}
For each evaluable stereogenic double bond, we compute the wrapped absolute
difference between predicted and experimental dihedral angles. Errors are
averaged within each ligand, then summarized by their mean, median, and 90th
percentile across ligands (degrees). A bond is evaluable when RDKit assigns
an \textit{E}/\textit{Z} configuration after hydrogen removal and both
double-bond atoms have a heavy-atom substituent, as in
Appendix~\ref{app:conformer-geometric-metrics}.

\textit{E}/\textit{Z} success requires every evaluable double bond in a ligand
to deviate by at most $10^\circ$ from the experimental reference;
\textit{R}/\textit{S} success requires every evaluable tetrahedral stereocenter
to match its reference assignment. A stereocenter is evaluable when its
\textit{R}/\textit{S} label is specified in the input and can be assigned in
the predicted pose. Success@1+stereo requires Success@1 and both
stereochemical criteria. 

\paragraph{Independent structural scoring.}
We score the same predictions with OpenStructure 2.12.0
(Table~\ref{tab:stereobench-ost}), reporting protein and interface metrics
independently of PoseBusters.

\begin{table}[!htbp]
\centering
\caption{\textbf{OpenStructure evaluation of the same predictions on StereoBench.} Computed with OpenStructure 2.12.0. The interface rows use the StereoBench multi-chain subset. OpenStructure contains no stereochemical term, so these scores are a preservation check on protein and interface quality. Best in \textbf{bold}.}
\label{tab:stereobench-ost}
\small
\setlength{\tabcolsep}{7pt}
\begin{tabular}{lccc}
\toprule
Metric & Vanilla & Boltz-steering & \textbf{AtomSteer} \\
\midrule
lDDT-PLI $\uparrow$ \ & 0.73 & \textbf{0.74} & 0.73 \\
\addlinespace
Complex lDDT $\uparrow$ \ & \textbf{0.95} & \textbf{0.95} & \textbf{0.95} \\
Backbone lDDT $\uparrow$ \ & \textbf{0.98} & \textbf{0.98} & \textbf{0.98} \\
TM-score $\uparrow$ \ & \textbf{0.98} & \textbf{0.98} & \textbf{0.98} \\
GDT-TS $\uparrow$ \  & \textbf{0.97} & \textbf{0.97} & \textbf{0.97} \\
C$\alpha$ RMSD (\AA{}) $\downarrow$ \ & \textbf{0.96} & \textbf{0.96} & 0.97 \\
\addlinespace
Interface lDDT $\uparrow$ \ & \textbf{0.77} & \textbf{0.77} & \textbf{0.77} \\
DockQ $\uparrow$ \ & \textbf{0.75} & \textbf{0.75} & \textbf{0.75} \\
\bottomrule
\end{tabular}
\end{table}

\paragraph{Evaluation subsets.}
For each metric, we include all targets with computable results for every
method shown in the table. Stereochemistry metrics additionally require
the corresponding evaluable \textit{E}/\textit{Z} bonds or
\textit{R}/\textit{S} centers.

\FloatBarrier
\subsection{Additional results}
\label{app:protein-ligand-additional-results}

\subsubsection{\textit{E}/\textit{Z} accuracy by bond count and threshold}
StereoBench contains 1{,}608 evaluable stereogenic double bonds across 1{,}103
ligands: 813 have one, 192 have two, and 98 have three or more.
Tables~\ref{tab:stereobench-strata} and~\ref{tab:stereobench-threshold} report accuracy
by bond count and angular threshold.
\begin{table}[!htbp]
\centering
\caption{\textbf{\textit{E}/\textit{Z} accuracy by number of stereogenic double bonds on StereoBench.} Best in \textbf{bold}, second-best \underline{underlined}.}
\label{tab:stereobench-strata}
\resizebox{\textwidth}{!}{%
\begin{tabular}{l r ccc ccc}
\toprule
& & \multicolumn{3}{c}{All \textit{E}/\textit{Z} bonds $\leq 10^\circ$ (\%) $\uparrow$}
  & \multicolumn{3}{c}{Median \textit{E}/\textit{Z} error ($^\circ$) $\downarrow$} \\
\cmidrule(lr){3-5}\cmidrule(lr){6-8}
Number of \textit{E}/\textit{Z} bonds & $N$ & Vanilla & Boltz-steering & AtomSteer & Vanilla & Boltz-steering & AtomSteer \\
\midrule
One bond        & 813 & 64.6 & \underline{66.2} & \textbf{81.2} & 5.63 & \underline{5.53} & \textbf{2.61} \\
Two bonds       & 192 & \underline{53.7} & 48.4 & \textbf{75.0} & \underline{6.46} & 6.96 & \textbf{2.83} \\
Three or more   &  98 & \underline{29.6} & 25.5 & \textbf{70.4} & 11.78 & \underline{8.06} & \textbf{2.19} \\
\bottomrule
\end{tabular}}
\end{table}
\begin{table}[!htbp]
\centering
\caption{\textbf{\textit{E}/\textit{Z} threshold sensitivity on StereoBench.} Percentage of the 1{,}103 ligands for which \emph{every} evaluable stereogenic double bond falls within the given threshold of the experimental ligand, the criterion of Table~\ref{tab:boltz-final} at $10^\circ$. Best per row in \textbf{bold}, second-best \underline{underlined}.}
\label{tab:stereobench-threshold}
\small
\setlength{\tabcolsep}{8pt}
\begin{tabular}{lccc}
\toprule
Threshold $\uparrow$ & Vanilla & Boltz-steering & \textbf{AtomSteer} \\
\midrule
$\leq2.5^\circ$ & 21.7 & \underline{21.8} & \textbf{47.1} \\
$\leq5^\circ$ & \underline{39.1} & \underline{39.1} & \textbf{63.4} \\
$\leq10^\circ$ & \underline{59.6} & 59.5 & \textbf{79.1} \\
$\leq15^\circ$ & 69.2 & \underline{70.8} & \textbf{85.9} \\
$\leq20^\circ$ & 73.3 & \underline{78.6} & \textbf{89.8} \\
$\leq30^\circ$ & 77.2 & \underline{92.4} & \textbf{93.0} \\
$\leq45^\circ$ & 79.6 & \textbf{96.0} & \underline{95.6} \\
\bottomrule
\end{tabular}
\end{table}

\subsubsection{Performance by PDB release date}
Splitting StereoBench by the Boltz-2 training cutoff (2023-06-01) gives 926
targets released before the cutoff and 177 on or after it; the corresponding
\textit{R}/\textit{S} subsets contain 451 and 87 targets.
Table~\ref{tab:stereobench-holdout} reports metrics recomputed within each split.

\begin{table}[!htbp]
\centering
\caption{\textbf{StereoBench split by PDB release date at the Boltz-2 training cutoff of 2023-06-01.} Same protocol, selector, and metric definitions as Table~\ref{tab:boltz-final}, recomputed within each split of the 1{,}103 targets. Best in \textbf{bold}, second-best \underline{underlined}.}
\label{tab:stereobench-holdout}
\small
\setlength{\tabcolsep}{5pt}
\begin{tabular}{llccc}
\toprule
Split & Metric & Vanilla & Boltz-steering & \textbf{AtomSteer} \\
\midrule
\multirow{6}{*}{\shortstack[l]{Before\\2023-06-01\\(N=926)}}
 & PB-valid $\uparrow$                        & 51.2 & \underline{90.6} & \textbf{92.5} \\
 & RMSD $<2$~\AA{} $\uparrow$                 & 67.9 & \underline{69.7} & \textbf{69.9} \\
 & Success@1 $\uparrow$                       & 38.1 & \underline{64.0} & \textbf{65.1} \\
 & \textit{E}/\textit{Z} median error ($^\circ$) $\downarrow$   & 6.02 & \underline{5.80} & \textbf{2.48} \\
 & All \textit{E}/\textit{Z} bonds $\leq10^\circ$ $\uparrow$    & \underline{61.0} & 60.7 & \textbf{80.5} \\
 & Success@1+stereo $\uparrow$                & 27.8 & \underline{39.8} & \textbf{49.7} \\
\midrule
\multirow{6}{*}{\shortstack[l]{On or after\\2023-06-01\\(N=177)}}
 & PB-valid $\uparrow$                        & 50.3 & \underline{97.2} & \textbf{97.7} \\
 & RMSD $<2$~\AA{} $\uparrow$                 & \textbf{38.4} & 37.3 & \underline{37.9} \\
 & Success@1 $\uparrow$                       & 23.2 & \underline{36.7} & \textbf{37.3} \\
 & \textit{E}/\textit{Z} median error ($^\circ$) $\downarrow$   & 8.17 & \underline{7.67} & \textbf{3.22} \\
 & All \textit{E}/\textit{Z} bonds $\leq10^\circ$ $\uparrow$    & 52.0 & \underline{53.1} & \textbf{72.3} \\
 & Success@1+stereo $\uparrow$                & 13.0 & \underline{19.2} & \textbf{22.0} \\
\bottomrule
\end{tabular}
\end{table}

\FloatBarrier
\subsubsection{Qualitative pose comparisons}
\begin{figure}[!htb]
    \centering
    \includegraphics[width=0.99\textwidth]{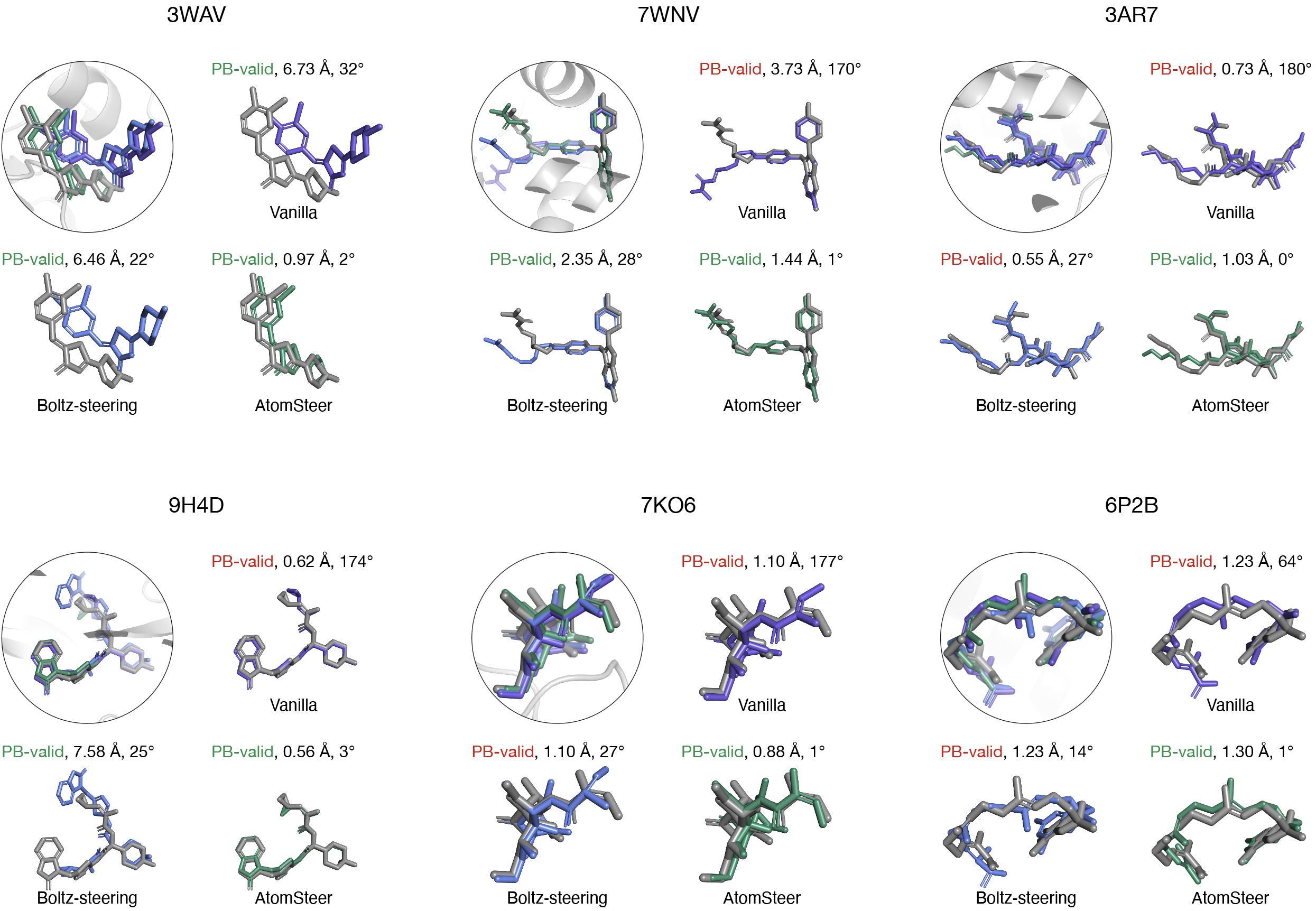}
    \caption{\textbf{Examples of AtomSteer correcting \textit{E}/\textit{Z}
    stereochemistry and pose validity.} Each panel overlays a confidence top-1
    ligand pose with the experimental ligand (gray), after protein alignment.
    PDB IDs identify the targets, and circular insets compare all three methods
    in the binding pocket. Annotations report PoseBusters validity
    (green = valid, red = invalid), ligand RMSD to the experimental reference
    (\AA), and \textit{E}/\textit{Z} dihedral error ($^\circ$). Methods shown:
    Vanilla (purple), Boltz-steering (blue), and AtomSteer (green).}
    \label{fig:boltz_asbetter}
\end{figure}

\begin{figure}[!htb]
    \centering
    \includegraphics[width=0.8\textwidth]{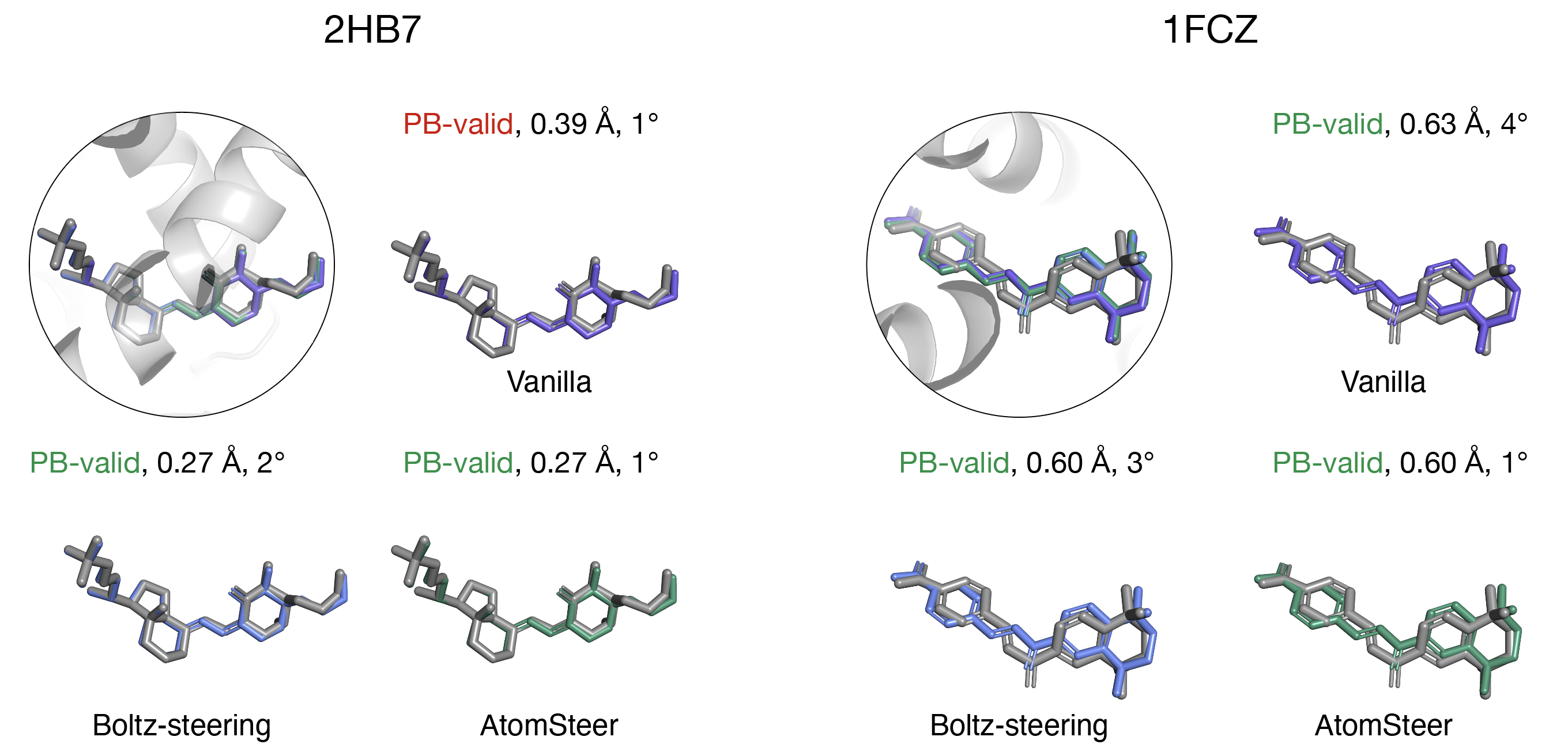}
    \caption{\textbf{Examples where AtomSteer preserves accurate
    \textit{E}/\textit{Z} geometry and ligand placement.} All methods have ligand
    RMSD $\leq 0.6$~\AA{}, and most poses are PB-valid. Layout and annotations
    follow Figure~\ref{fig:boltz_asbetter}.} 
\label{fig:boltz_asequal}
\end{figure}

\begin{figure}[!htb]
    \centering
    \includegraphics[width=0.8\textwidth]{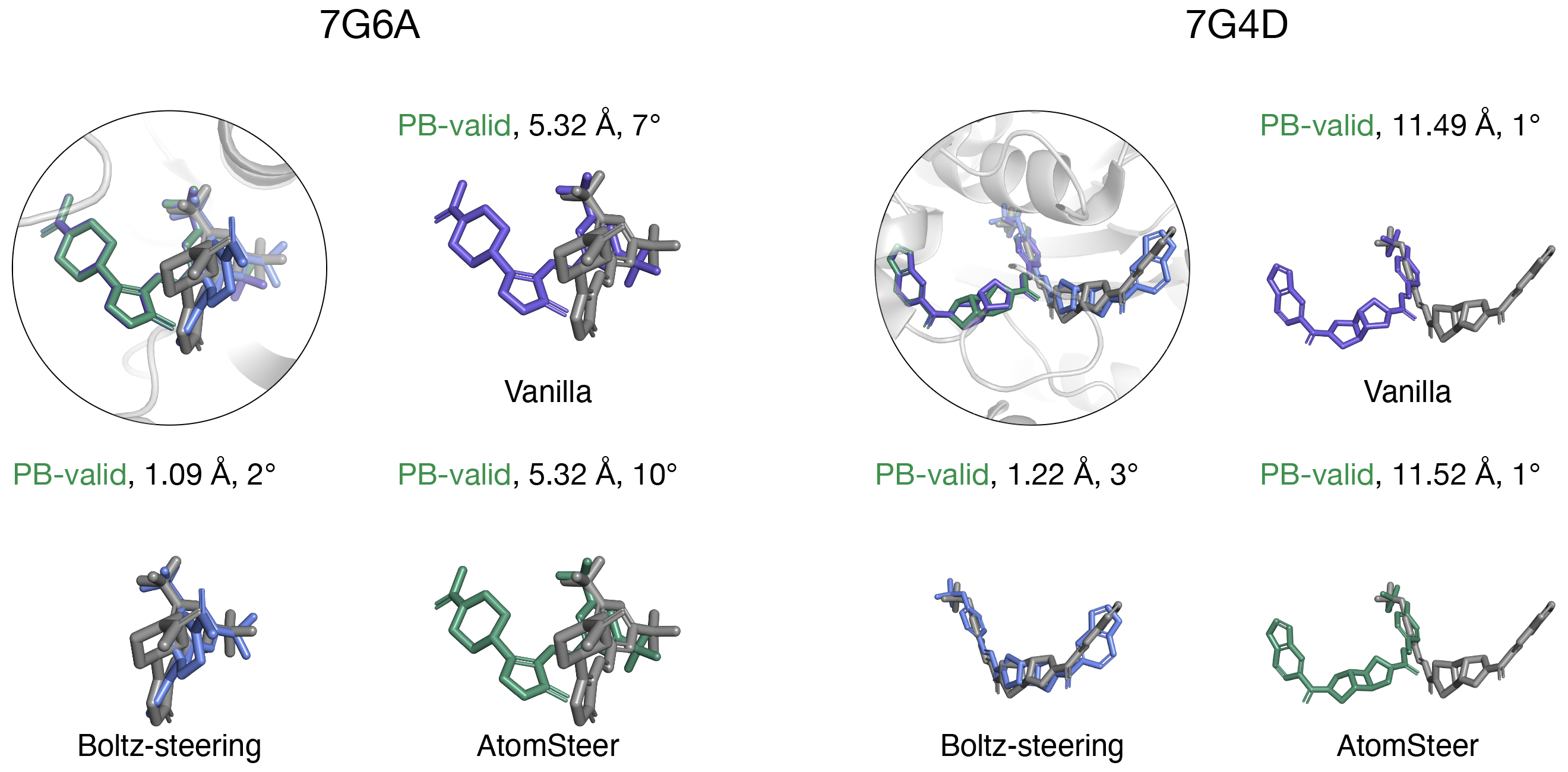}
    \caption{\textbf{Examples where Boltz-steering achieves lower ligand
    RMSD than AtomSteer.} All methods have \textit{E}/\textit{Z} dihedral errors
    within $10^\circ$. For 7G4D, ligand RMSDs are 1.2~\AA{} (Boltz-steering)
    and 11.5~\AA{} (AtomSteer); for 7G6A, they are 1.1~\AA{} and 5.3~\AA{},
    respectively. Layout and annotations follow Figure~\ref{fig:boltz_asbetter}.}
    \label{fig:boltz_asworse}
\end{figure}

\FloatBarrier

\end{document}